\documentclass[10pt,twocolumn,letterpaper]{article}

\usepackage[pagenumbers]{cvpr} 
\usepackage{stfloats}
\usepackage{graphicx}
\usepackage{amsmath}
\usepackage{amssymb}
\usepackage{booktabs}
\usepackage{multirow}
\usepackage{bbm}
\usepackage{amsmath}
\usepackage{array}
\usepackage{enumitem}
\usepackage{makecell}
\usepackage{float}
\usepackage{longtable}
\usepackage{rotating}
\usepackage{lipsum}
\DeclareMathOperator*{\argmax}{arg\,max}

\newcolumntype{x}[1]{%
>{\centering\hspace{0pt}}p{#1}}%

\setlist{noitemsep,topsep=3pt} 

\usepackage[pagebackref,breaklinks,colorlinks]{hyperref}

\usepackage[capitalize]{cleveref}
\crefname{section}{Sec.}{Secs.}
\Crefname{section}{Section}{Sections}
\Crefname{table}{Table}{Tables}
\crefname{table}{Tab.}{Tabs.}

\def\confName{CVPR}
\def\confYear{2023}

\begin{document}

\title{\vspace{-2em} PartSLIP: Low-Shot Part Segmentation for 3D Point Clouds via Pretrained Image-Language Models}

\author{Minghua Liu\textsuperscript{1}\quad Yinhao Zhu\textsuperscript{2}\quad Hong Cai\textsuperscript{2}\quad Shizhong Han\textsuperscript{2}\quad Zhan Ling\textsuperscript{1}\quad Fatih Porikli\textsuperscript{2}\quad Hao Su\textsuperscript{1}\\
\textsuperscript{1}UC San Diego \quad \textsuperscript{2}Qualcomm AI Research\thanks{Qualcomm AI Research is an initiative of Qualcomm Technologies, Inc.}
\\
\\
Project Website: \url{https://colin97.github.io/PartSLIP_page/}
}

  
\makeatletter
\let\@oldmaketitle\@maketitle
\renewcommand{\@maketitle}{\@oldmaketitle
    \vspace{-1.5\baselineskip}
    \includegraphics[width=\linewidth]{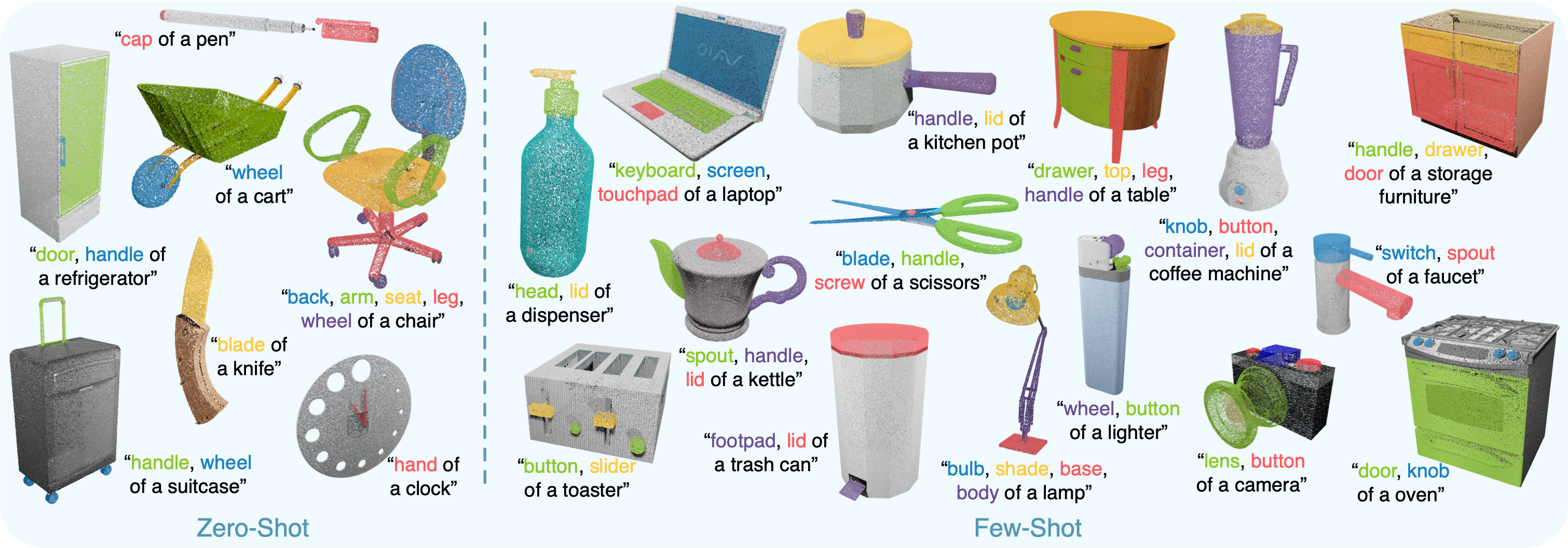}
    \captionof{figure}{We propose PartSLIP, a zero/few-shot method for 3D point cloud part segmentation by leveraging pretrained image-language models. The figure shows text prompts and corresponding semantic segmentation results (zoom in for details). Our method also supports part-level instance segmentation. See Figure~\ref{fig:instance_seg} and Figure~\ref{fig:real_world_demo} for more results.}
    \label{fig:teaser}
\bigskip}
\makeatother

\maketitle


\vspace{-1em}
\begin{abstract}
\vspace{-1em}
Generalizable 3D part segmentation is important but challenging in vision and robotics. Training deep models via conventional supervised methods requires large-scale 3D datasets with fine-grained part annotations, which are costly to collect. This paper explores an alternative way for low-shot part segmentation of 3D point clouds by leveraging a pretrained image-language model, GLIP, which achieves superior performance on open-vocabulary 2D detection. We transfer the rich knowledge from 2D to 3D through GLIP-based part detection on point cloud rendering and a novel 2D-to-3D label lifting algorithm. We also utilize multi-view 3D priors and few-shot prompt tuning to boost performance significantly. Extensive evaluation on PartNet and PartNet-Mobility datasets shows that our method enables excellent zero-shot 3D part segmentation. \textbf{Our few-shot version not only outperforms existing few-shot approaches by a large margin but also achieves highly competitive results compared to the fully supervised counterpart.} Furthermore, we demonstrate that our method can be directly applied to iPhone-scanned point clouds without significant domain gaps.
\vspace{-12pt}
\end{abstract}
\vspace{-0.7em}
\section{Introduction}
\vspace{-0.7em}
\label{sec:introduction}

Human visual perception can parse objects into parts and generalize to unseen objects, which is crucial for understanding their structure, semantics, mobility, and functionality. 3D part segmentation plays a critical role in empowering machines with such ability and facilitates a wide range of applications, such as robotic manipulation, AR/VR, and shape analysis and synthesis~\cite{aleotti20123d,liu2022frame,xu2022unsupervised,mo2019structurenet}. 

Recent part-annotated 3D shape datasets~\cite{mo2019partnet,yi2016scalable,xiang2020sapien} have promoted advances in designing various data-driven approaches for 3D part segmentation~\cite{qian2022pointnext,wang2019dynamic,liu2019relation,yi2019gspn}. While standard supervised training enables these methods to achieve remarkable results, they often struggle with out-of-distribution test shapes (e.g., unseen classes). However, compared to image datasets, these 3D part-annotated datasets are still orders of magnitude smaller in scale, since building 3D models and annotating fine-grained 3D object parts are laborious and time-consuming. It is thus challenging to provide sufficient training data covering all object categories. For example, the recent PartNet dataset ~\cite{mo2019partnet} contains only 24 object categories, far less than what an intelligent agent would encounter in the real world.

To design a generalizable 3D part segmentation module, many recent works have focused on the few-shot setting, assuming only a few 3D shapes of each category during training. They design various strategies to learn better representations, and complement vanilla supervised learning~\cite{zhao2021few,liu2022autogpart,sun2022semi,wang2020few,sharma2022mvdecor}. While they show improvements over the original pipeline, there is still a large gap between what these models can do and what downstream applications need. The problem of generalizable 3D part segmentation is still far from being solved. Another parallel line of work focuses on learning the concept of universal object parts and decomposing a 3D shape into a set of (hierarchical) fine-grained parts~\cite{wang2021learning,luo2020learning,yu2019partnet}. However, these works do not consider the semantic labeling of parts and may be limited in practical use.

In this paper, we seek to solve the low-shot (zero- and few-shot) 3D part segmentation problem by leveraging pretrained image-language models, inspired by their recent striking performances in low-shot learning. By pretraining on large-scale image-text pairs, image-language models~\cite{radford2021learning,jia2021scaling,li2022grounded,zhang2022glipv2,alayrac2022flamingo,ramesh2022hierarchical,saharia2022photorealistic} learn a wide range of visual concepts and knowledge, which can be referenced by natural language. Thanks to their impressive zero-shot capabilities, they have already enabled a variety of 2D/3D vision and language tasks~\cite{hong2022avatarclip,sanghi2022clip,zhang2022pointclip,jain2021putting,corona2022voxel,rozenberszki2022language,rao2022denseclip}.

As shown in Figure~\ref{fig:teaser}, our method takes a 3D point cloud and a text prompt as input, and generates both 3D semantic and instance segmentations in a zero-shot or few-shot fashion. Specifically, we integrate the GLIP~\cite{li2022grounded} model, which is pretrained on 2D visual grounding and detection tasks with over 27M image-text pairs and has a strong capability to recognize object parts. To connect our 3D input with the 2D GLIP model, we render multi-view 2D images for the point cloud, which are then fed into the GLIP model together with a text prompt containing part names of interest. The GLIP model then detects parts of interest for each 2D view and outputs detection results in the form of 2D bounding boxes. Since it is non-trivial to convert 2D boxes back to 3D, we propose a novel 3D voting and grouping module to fuse the multi-view 2D bounding boxes and generate 3D instance segmentation for the input point cloud. Also, the pretrained GLIP model may not fully understand our definition of parts only through text prompts. We find that an effective solution is prompt tuning with few-shot segmented 3D shapes. In prompt tuning, we learn an offset feature vector for the language embedding of each part name while fixing the parameters of the pretrained GLIP model. Moreover, we propose a multi-view visual feature aggregation module to fuse the information of multiple 2D views, so that the GLIP model can have a better global understanding of the input 3D shape instead of predicting bounding boxes from each isolated 2D view.

To better understand the generalizability of various approaches and their performances in low-shot settings, we propose a benchmark PartNet-Ensembled (PartNetE) by incorporating two existing datasets PartNet~\cite{mo2019partnet} and PartNetMobility~\cite{xiang2020sapien}. Through extensive evaluation on PartNetE, we show that our method enables excellent zero-shot 3D part segmentation. With few-shot prompt tuning, our method not only outperforms existing few-shot approaches by a large margin but also achieves highly competitive performance compared to the fully supervised counterpart. We also demonstrate that our method can be directly applied to iPhone-scanned point clouds without significant domain gaps. In summary, our contributions mainly include:

\vspace{-0.2em}
\begin{itemize}[leftmargin=*] 
\item We introduce a novel 3D part segmentation method that leverages pretrained image-language models and achieves outstanding zero-shot and few-shot performance.
\item We present a 3D voting and grouping module, which effectively converts multi-view 2D bounding boxes into 3D semantic and instance segmentation.
\item We utilize few-shot prompt tuning and multi-view feature aggregation to boost GLIP's detection performance.
\item We propose a benchmark PartNetE that benefits future work on low-shot and text-driven 3D part segmentation.
\end{itemize}  
\section{Related Work}
\label{sec:related_work}

\begin{figure*}[t]
    \centering
    \includegraphics[width=\textwidth]{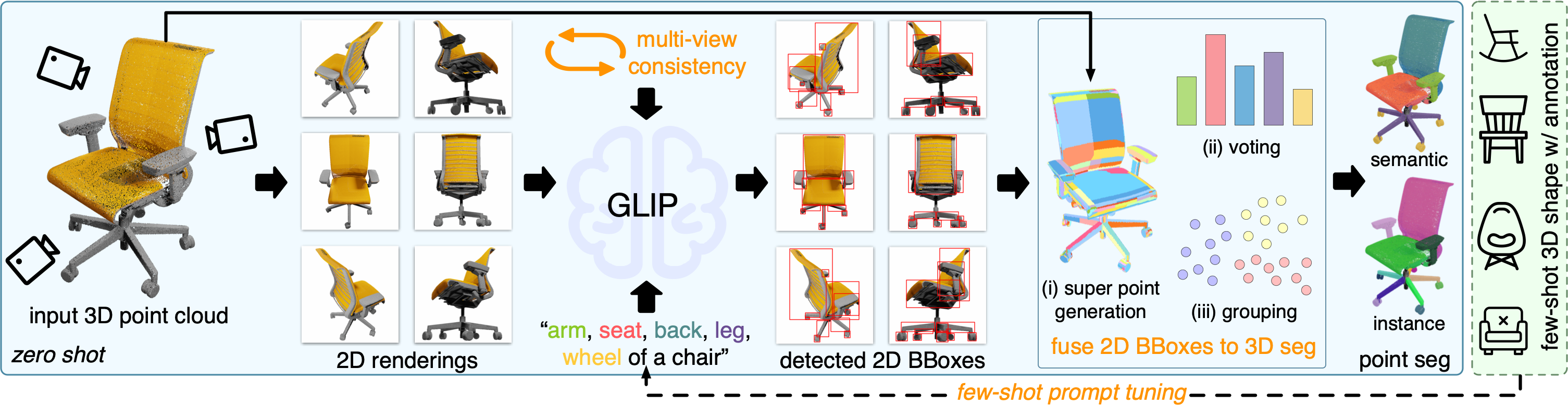}
    \caption{The figure shows our overall pipeline. Our proposed components are highlighted in orange.}
    \label{fig:overview}
    \vspace{-1em}
\end{figure*}

\subsection{3D Part Segmentation}
\vspace{-0.5em}
3D part segmentation involves two main tasks: semantic segmentation and instance segmentation. Most 3D backbone networks~\cite{qi2017pointnet++,qian2022pointnext,wang2019dynamic,thomas2019kpconv} are capable of semantic segmentation by predicting a semantic label for each geometric primitive (e.g., point or voxel). Existing learning-based approaches solve instance segmentation by incorporating various grouping~\cite{jiang2020pointgroup,liu2020self,wang2018sgpn,zhang2021point,he2020learning,vu2022softgroup,wang2019associatively,chu2021icm} or region proposal~\cite{yi2019gspn,yang2019learning,hou20193d} strategies into the pipeline. Different from standard training with per-point part labels, some works leverage weak supervision, such as bounding box~\cite{liu2022box2seg,chibane2022box2mask}, language reference game~\cite{koo2022partglot}, or IKEA manual~\cite{wangikea}. Instead of focusing on single objects, \cite{bokhovkin2021towards,notchenko2022scan2part} also consider part segmentation for scene-scale input. Moreover, unlike the two classical tasks of semantic and instance segmentation, another parallel line of works decomposes a 3D shape into a set of (hierarchical) fine-grained parts but without considering semantic labels~\cite{wang2021learning,luo2020learning,yu2019partnet}, which differs from our objective. Recently, some works also propose to learn a continuous implicit semantic field~\cite{kohli2020semantic,zhi2021place}. 

\vspace{-0.3em}
\subsection{Data-Efficient 3D Segmentation}
\vspace{-0.5em}
In order to train a generalizable 3D part segmentation network with low-shot data, many existing efforts focus on leveraging various pretext tasks and auxiliary losses~\cite{gadelha2020label,sharma2022prifit,hassani2019unsupervised,thabet2019mortonnet,alliegro2021joint}. In addition, \cite{naeem20223d,han2020compositionally} studies the compositional generalization of 3D parts. \cite{wang2020few} deforms input shapes to align with few-shot template shapes. \cite{sharma2022mvdecor} leverages 2D contrastive learning by projecting 3D shapes and learning dense multi-view correspondences. \cite{chen2019bae} leverages branched autoencoders to co-segment a collection of shapes. Also, some works aim to learn better representations by utilizing prototype learning~\cite{zhao2021few}, reinforcement learning~\cite{liu2022autogpart}, and data augmentation~\cite{sun2022semi}. Moreover, there is a line of work investigating label-efficient 3D segmentation~\cite{xu2020weakly,hou2021exploring,liu2022less,liu2021one,zhang2021weakly,zhang2021perturbed,zhang2021weakly,yang2022mil}, assuming a small portion of training data is annotated (e.g., 0.1\% point labels). While the setting may be useful in indoor and autonomous driving scenarios, it is not aligned with our goal since the number of training shapes is already limited in our setup.

\vspace{-0.3em}
\subsection{3D Learning with Image-Language Models}
\vspace{-0.5em}
Pretrained image-language models have recently made great strides by pretraining on large-scale image-text pairs~\cite{radford2021learning,jia2021scaling,li2022grounded,zhang2022glipv2,alayrac2022flamingo,ramesh2022hierarchical,saharia2022photorealistic}. Due to their learned rich visual concepts and impressive zero-shot capabilities, they have been applied to a wide range of 3D vision tasks, such as 3D avatar generation and manipulation~\cite{hong2022avatarclip,canfes2022text,jetchev2021clipmatrix}, general 3D shape generation~\cite{sanghi2022clip,michel2022text2mesh,khalid2022text,jain2022zero}, low-shot 3D shape classification~\cite{zhang2022pointclip}, neural radiance fields~\cite{wang2022clip,jain2021putting}, 3D visual grounding~\cite{thomason2022language,corona2022voxel}, and 3D representation learning~\cite{rozenberszki2022language}. To the best of our knowledge, we are one of the first to utilize pretrained image-language models to help with the task of 3D part segmentation.

\vspace{-0.5em}
\section{Proposed Method: PartSLIP}
\label{sec:method}

\vspace{-0.2em}
\subsection{Overview: 3D Part Segmentation with GLIP}
\vspace{-0.2em}

We aim to solve both semantic and instance segmentation for 3D object parts by leveraging pretrained image-language models (ILMs). There are various large-scale ILMs emerged in the past few years. In order to enable generalizable 3D object part segmentation, the pre-trained ILM is expected to be capable of generating region-level output (e.g., 2D segmentation or 2D bounding boxes) and recognizing object parts. After comparing several released pretrained ILMs (e.g., CLIP~\cite{radford2021learning}), we find that the GLIP~\cite{li2022grounded} model is a good choice. The GLIP~\cite{li2022grounded} model focuses on 2D visual grounding and detection tasks. It takes as input a free-form text description and a 2D image, and locates all phrases of the text by outputting multiple 2D bounding boxes for the input image. By pretraining on large-scale image-text pairs (e.g., 27M grounding data), the GLIP model learns a wide range of visual concepts (e.g., object parts) and enables open-vocabulary 2D detection.

Figure~\ref{fig:overview} shows our overall pipeline, where we take a 3D point cloud as input. Here, we consider point clouds from unprojecting and fusing multiple RGB-D images, which is a common setup in real-world applications and leads to dense points with color and normal. To connect the 2D GLIP model with our 3D point cloud input, we render the point cloud from $K$ predefined camera poses. The camera poses are uniformly spaced around the input point cloud, aiming to cover all regions of the shape. Since we assume a dense and colored point cloud input\footnote{Recent commodity-grade 3D scanning devices (e.g., iPhone 12 Pro) can already capture high-quality point clouds (see Figure~\ref{fig:real_world_demo}).}, we render the point cloud by simple rasterization without introducing significant artifacts. The $K$ rendered images are then fed separately into the pretrained GLIP model along with a text prompt. We format the text prompt by concatenating all part names of interest and the object category. For example, for a chair point cloud, the text prompt could be ``arm, back, seat, leg, wheel of a chair''. Please note that unlike the traditional segmentation networks, which are limited to a closed set of part categories, our method is more flexible and can include any part name in the text prompt. For each 2D rendered image, the GLIP model is expected to predict multiple bounding boxes, based on the text prompt, for all part instances that appear. We then fuse all bounding boxes from $K$ views into 3D to generate semantic and instance segmentation for the input point cloud (Section~\ref{sec:2dbbox_to_3dseg}).

The above pipeline introduces an intuitive zero-shot approach for 3D part segmentation without requiring any 3D training. However, its performance may be limited by the GLIP predictions. We thus propose two additional components, which could be incorporated into the above pipeline to encourage more accurate GLIP prediction: (a) prompt tuning with few-shot 3D data, which enables the GLIP model to quickly adapt to the meaning of each part name (Section~\ref{sec:prompt_tuning}); (b) multi-view feature aggregation, which allows the GLIP model to have a more comprehensive visual understanding of the input 3D shape (Section~\ref{sec:multi_view_feature_aggregation}).

\vspace{-0.2em}
\subsection{Detected 2D BBoxes to 3D Point Segmentation}
\vspace{-0.2em}
\label{sec:2dbbox_to_3dseg}

Although the correspondence between 2D pixels and 3D points are available, there are still two main challenges when converting the detected 2D bounding boxes to 3D point segmentation. First, bounding boxes are not as precise as point-wise labels. A 2D bounding box may cover points from other part instances as well. Also, although each bounding box may indicate a part instance, we are not provided with their relations across views. It's not very straightforward to determine which sets of 2D bounding boxes indicate the same 3D part instance.

Therefore, we propose a learning-free module to convert the GLIP predictions to 3D point segmentation, which mainly includes three steps: (a) oversegment the input point cloud into a collection of super points; (b) assign a semantic label for each super point by 3D voting; and (c) group super points within each part category into instances based on their similarity of bounding box coverage.

\noindent\textbf{3D Super Point Generation:} We follow the method in~\cite{landrieu2018large} to oversegment the input point cloud into a collection of super points. Specifically, we utilize point normal and color as features and solve a \textit{generalized minimal partition problem} with an $l_0$-cut pursuit algorithm~\cite{landrieu2017cut}. Since points in each generated super point share similar geometry and appearance, we assume they belong to one part instance. The super point partition serves as an important 3D prior when assigning semantic and instance labels. It also speeds up the label assignment, as the number of super points is orders of magnitude smaller than the number of 3D points.

\noindent\textbf{3D Semantic Voting:} While a single bounding box may cover irrelevant points from other parts, we want to leverage information from multiple views and the super point partition to counteract the effect of irrelevant points. Specifically, for each pair of super point and part category, we calculate a score $s_{i,j}$ measuring the proportion of the $i$th super point covered by any bounding box of part category $j$:
\vspace{-0.5em}
\begin{equation}\label{eq:semseg_score}
s_{i,\,j} = \frac{\sum_k \sum_{p \in SP_i} [\operatorname{VIS}_k(p)] [\exists b\in BB_k^j : \operatorname{INS}_b(p)]}{\sum_k \sum_{p \in SP_i} [\operatorname{VIS}_k(p)]},
\vspace{-0.5em}
\end{equation}
where $SP_i$ indicates the $i$th super point, $[\cdot]$ is the Iverson bracket, $\operatorname{VIS}_k(p)$ indicates whether the 3D point $p$ is visible in view $k$, $BB_k^j$ is a list of predicted bounding boxes of category $j$ in view $k$, and $\operatorname{INS}_b(p)$ indicates whether the projection of point $p$ in view $k$ is inside the bounding box $b$. 

Note that for each view, we only consider visible points since bounding boxes only contain visible portions of each part instance. Both $\operatorname{VIS}_k(p)$ and $ \operatorname{INS}_b(p)$ can be computed based on the information from point cloud rasterization. After that, for each super point $i$, we assign part category $j$ with the highest score $s_{i,j}$ to be its semantic label.

\noindent\textbf{3D Instance Grouping:} In order to group the super points into part instances, we first regard each super point as an individual instance and then consider whether to merge each pair of super points. For a pair of super points $SP_u$ and $SP_v$, we merge them if: (a) they have the same semantic label, (b) they are adjacent in 3D,  and (c) for each bounding box, they are either both included or both excluded. 

Specifically, for the second criterion, we construct a kNN graph for the input points to check whether $SP_u$ and $SP_v$ are adjacent in 3D. For the third criterion, we consider bounding boxes from views where both of them are visible: 
\vspace{-0.8em}
\begin{equation}
    B = \{b \in BB_k | \operatorname{VIS}_k(SP_u) \wedge \operatorname{VIS}_k(SP_v)\},
\vspace{-0.2em}
\end{equation} 
where $\operatorname{VIS}_k(SP_u)$ indicates whether the super point $SP_u$ can be (partially) visible in view $k$ and $BB_k$ indicates all predicted bounding boxes of view $k$. Suppose $B$ contains $n$ bounding boxes. We then construct two $n$ dimensional vectors $I_u$ and $I_v$, describing the bounding box coverage of $SP_u$ and $SP_v$. Specifically, $I_u[i]$ is calculated as:
\vspace{-0.5em}
\begin{equation}
  I_u[i] = \frac{\sum_{p\in SP_u} [\operatorname{VIS}_{B[i]}(p)][\operatorname{INS}_{B[i]}(p)]}{\sum_{p\in SP_u}[\operatorname{VIS}_{B[i]}(p)]},
\vspace{-0.5em}
\end{equation} where $B[i]$ indicates the $i$th bounding box of $B$, $\operatorname{VIS}_{B[i]}(p)$ indicates whether $p$ is visible in the corresponding view of $B[i]$, and $\operatorname{INS}_{B[i]}(p)$ indicates whether the projection of $p$ is inside $B[i]$. If $\frac{|I_u-I_v|_1}{max(|I_u|_1, |I_v|_1)}$ is smaller then a predefined threshold $\tau$, we consider they satisfy the third criterion.

After checking all pairs of super points, the super points are divided into multiple connected components, each of which is then considered to be a part instance. We found that our super point-based module works well in practice.

\vspace{-0.2em}
\subsection{Prompt Tuning w/ Few-Shot 3D Data}
\vspace{-0.2em}
\label{sec:prompt_tuning}

In our method, we utilize natural language to refer to a part. However, natural language can be flexible. An object part can be named in multiple ways (e.g., spout and mouth for kettles; caster and wheel for chairs), and the definition of some parts may be ambiguous (see the dispenser in Figure~\ref{fig:teaser}). We thus hope to finetune the GLIP model using a few 3D shapes with ground truth part segmentation, so that the GLIP model can quickly adapt to the actual definition of the part names in the text prompt.

\begin{figure}[t]
    \centering
    \includegraphics[width=\linewidth]{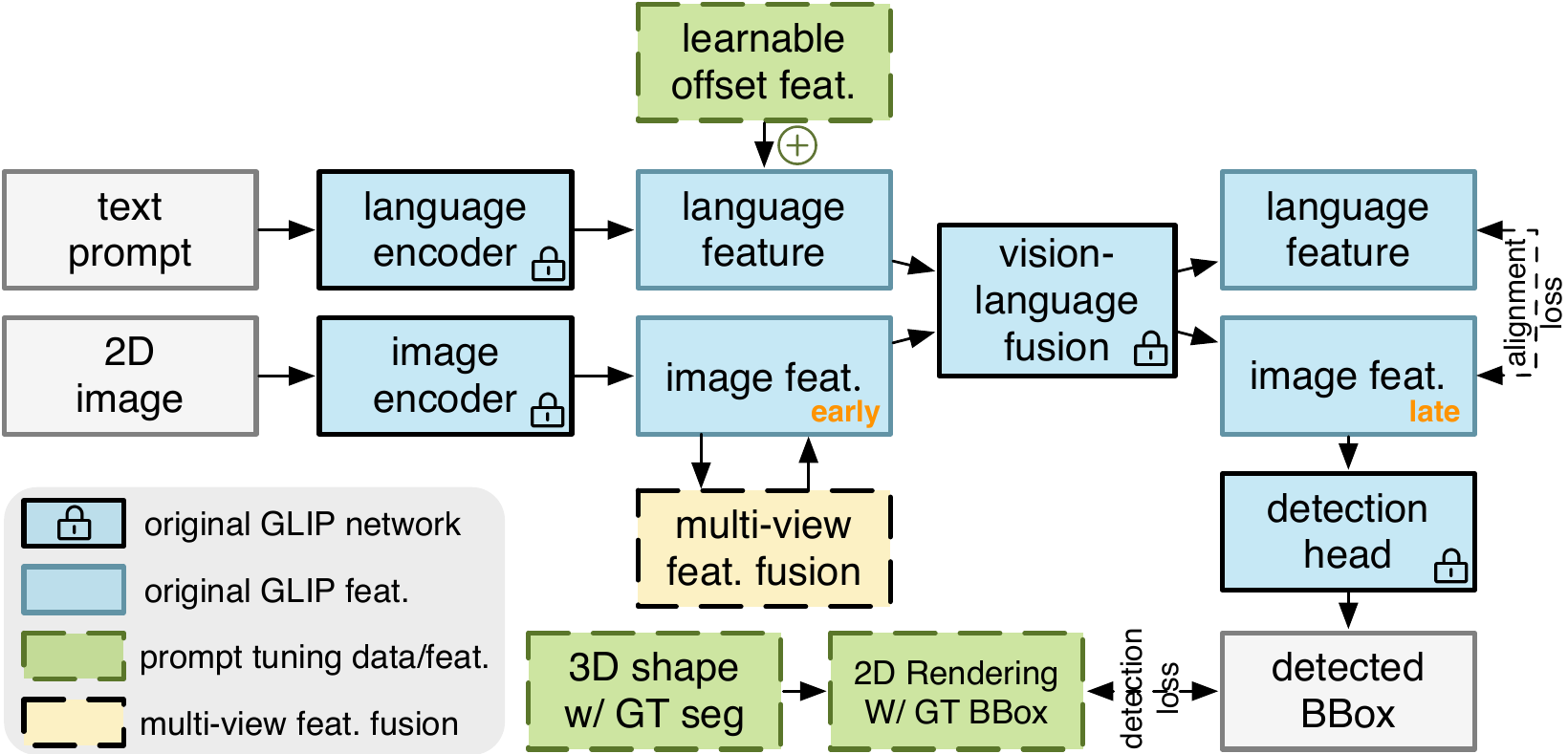}
    \caption{The original GLIP pipeline and our additional modules: few-shot prompt tuning and multi-view feature aggregation. We find that early fusion leads to better performance than late fusion.}
    \label{fig:glip_pipeline}
    \vspace{1em}
\end{figure}

Figure~\ref{fig:glip_pipeline} shows the overall architecture of the GLIP model. It first employs a language encoder and an image encoder to extract language features and multi-scale visual features, respectively, which are then fed into a vision-language fusion module to  fuse information across modalities. The detection head then takes as input the language-aware image features and predicts 2D bounding boxes. During pretraining, the GLIP network is supervised by both detection loss and image-language alignment loss.

It is not desirable to change the parameters of the visual module or the entire GLIP model since our goal is to leverage only a few 3D shapes for finetuning. Instead, we follow the prompt tuning strategy introduced in GLIP~\cite{li2022grounded} to finetune only the language embedding of each part name while freezing the parameters of the pretrained GLIP model. Specifically, we perform prompt tuning for each object category separately. Suppose the input text of an object category includes $l$ tokens and denote the extracted language features (before VL fusion) as $f_l \in \mathbb{R}^{l\times c}$, where $c$ is the number of channels. We aim to learn offset features $f_o \in \mathbb{R}^{l\times c}$ for $f_l$ and feed their summation $f_l + f_o$ to the remaining GLIP pipeline. The offset features $f_o$ consist of constant vectors for each token (part name), which can be interpreted as a local adjustment of the part definition in the language embedding space. Note that $f_o$ is not predicted by a network but is directly optimized as a trainable variable during prompt tuning. Also, $f_o$ will be fixed for each object category after prompt tuning.

In order to utilize the detection and alignment losses for optimization, we convert the few-shot 3D shapes with ground truth instance segmentation into 2D images with bounding boxes. Specifically, for each 3D point cloud, we render $K$ 2D images from the predefined camera poses. For generating corresponding 2D ground-truth bounding boxes, we project each part instance from 3D to 2D. Note that, after projection, we need to remove occluded points (i.e., invisible points of each view) and noisy points (i.e., visible but isolated in tiny regions) to generate reasonable bounding boxes. We find that by prompt tuning with only one or a few 3D shapes, the GLIP model can quickly adapt to our part definitions and generalize to other instances.

\vspace{-0.5em}
\subsection{Multi-View Visual Feature Aggregation}
\vspace{-0.5em}
\label{sec:multi_view_feature_aggregation}

The GLIP model is sensitive to camera views. For example, images taken from some unfamiliar views (e.g., the rear view of a cabinet) can be uninformative and confusing, making it difficult for the GLIP model to predict accurately. However, unlike regular 2D recognition tasks, our input is a 3D point cloud, and there are pixel-wise correspondences between different 2D views. Therefore, we hope the GLIP model can leverage these 3D priors to make better predictions instead of focusing on each view in isolation.

In order to take full advantage of the pretrained GLIP model, we propose a training-free multi-view visual feature aggregation module that could be plugged into the original GLIP network without changing any existing network weights. Specifically, the feature aggregation module takes $K$ feature maps $\{f_k \in \mathbb{R}^{m\times m\times c} \}$ as input, where $m$ is the spatial resolution of the feature map and $c$ is the number of channels. The input feature maps $\{f_k\}$ are generated by the GLIP module separately for each 2D view of the input point cloud. Our feature aggregation module fuses them and generates $K$ fused feature maps $\{f_k^{\prime} \}$ of the same shape, which are then used to replace the original feature maps and fed into the remaining layers of the GLIP model.

\begin{figure}[t]
    \centering
    \includegraphics[width=\linewidth]{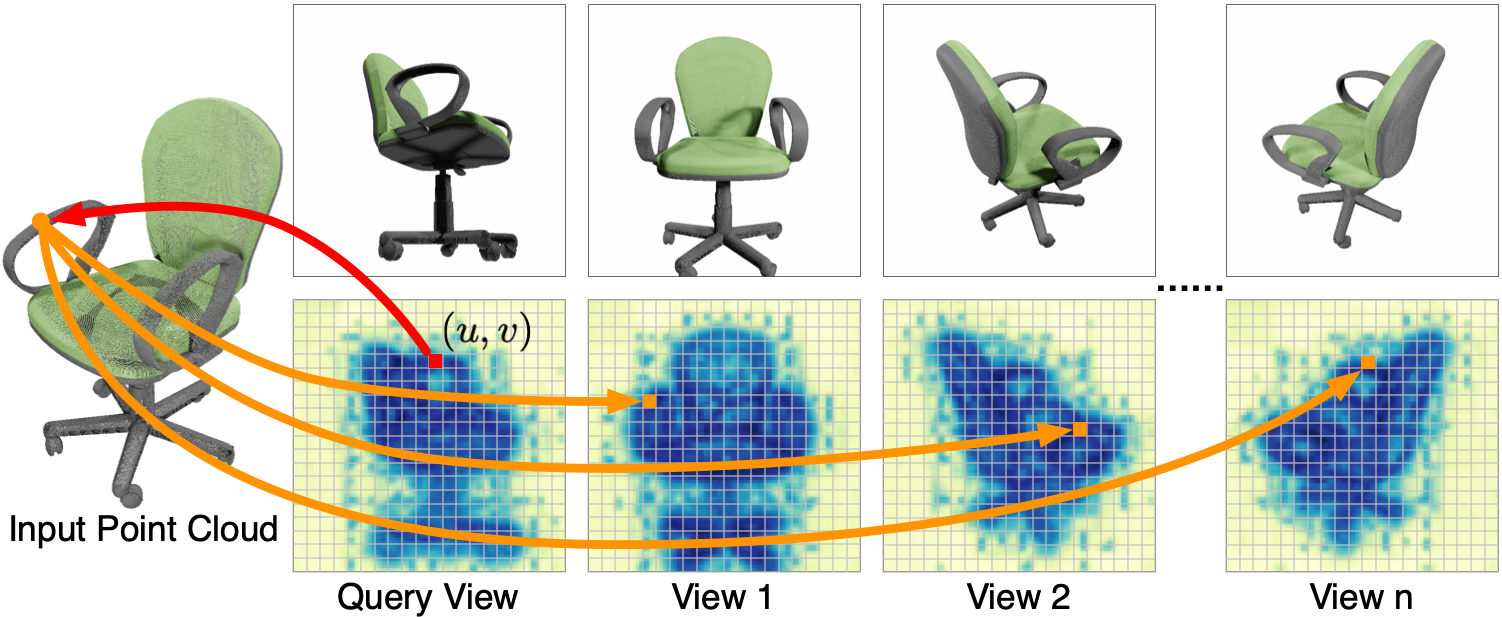}
    \caption{Multi-view 2D renderings (first row) and their feature maps (second row). For a feature cell (red), we aggregate all its corresponding feature cells (orange) across views.}
    \label{fig:multi-view-feat-aggregation}
    \vspace{1em}
\end{figure}

As shown in Figure~\ref{fig:multi-view-feat-aggregation}, for each cell $(u, v)$ of feature map $f_i$, we find its corresponding cell $(u^{i\rightarrow k}, v^{i\rightarrow k})$ in each feature map $f_k$ and use their weighted average to serve as the fused feature of the cell:
\vspace{-0.5em}
\begin{equation}
f_i^{\prime}[u,v] = \frac{1}{\sum_k w_{u,v}^{i\rightarrow k}} \sum_k w_{u,v}^{i\rightarrow k} f_k[u^{i\rightarrow k}, v^{i\rightarrow k}]. 
\vspace{-0.5em}
\end{equation}
Specifically, we define $P_i(u,v)$ as the set of 3D points that are visible in view $i$ and whose projections lie within cell $(u,v)$. We then choose the cell in view $k$ with the most overlapping 3D points as the corresponding cell: $(u^{i\rightarrow k}, v^{i\rightarrow k}) = \argmax\limits_{(x,y)} |P_i(u,v) \cap P_k(x,y)|$ and define the weights $w_{u,v}^{i\rightarrow k}$ as $\frac{|P_i(u,v) \cap P_k(u^{i\rightarrow k}, v^{i\rightarrow k})|}{|P_i(u,v)|}$. Note that if all 3D points in $P_i(u,v)$ are not visible in a view $k$, then feature map $f_k$ will not contribute to $f_i^{\prime}[u,v]$. Since the GLIP model generates multi-scale visual features, our aggregation module fuses features of each scale level separately. 

There are various options for which visual features to fuse (see Figure~\ref{fig:glip_pipeline}). One intuitive choice is to fuse the final visual features before the detection head, and we denote this choice as \emph{late fusion}. We find that the late fusion does not improve or even degrade the original performance. This is mainly because the final visual features contain too much shape information of the predicted 2D bounding boxes. Directly averaging the final visual features can somehow be seen as averaging bounding boxes in 2D, which does not make sense. Instead, we choose to fuse the visual features before the vision-language fusion (denoted as \emph{early fusion}). Since the text prompt is not involved yet, the visual features mainly describe the geometry and appearance of the input shape. Fusing these features across views with the 3D priors can thus lead to a more comprehensive visual understanding of the input shape.
\section{Experiments}
\label{sec:experiments}
\vspace{-0.5em}
\subsection{Datasets and Metrics}
\vspace{-0.5em}

To evaluate the generalizability of various approaches and their performances in the low-shot setting, we curate an ensembled dataset named PartNet-Ensembled (PartNetE), which consists of shapes from existing datasets PartNet~\cite{mo2019partnet} and PartNet-Mobility~\cite{xiang2020sapien}. Note that PartNet-Mobility contains more object categories but fewer shape instances, and PartNet contains more shape instances but fewer object categories. We thus utilize shapes from PartNet-Mobility for few-shot learning and test, and use shapes from PartNet to serve as additional large-scale training data for transfer learning. As a result, the test set of PartNetE contains 1,906 shapes covering 45 object categories. In addition, we randomly reserve 8 shapes from each of the 45 object categories for few-shot training. Also, we may utilize the additional 28,367 shapes from PartNet for training, which cover 17 out of 45 object categories and have consistent part annotations as the test set. Some of the original part categories in PartNet (e.g., ``back\_frame\_vertical\_bar" for chairs) are too fine-grained and ambiguous to evaluate unsupervised text-driven part segmentation approaches. We thus select a subset of 103 parts when constructing the PartNetE dataset, which covers both common coarse-grained parts (e.g., chair back and tabletop) and fine-grained parts (e.g., wheel, handle, button, knob, switch, touchpad) that may be useful in downstream tasks such as robotic manipulation. See supplementary for more details of the dataset.

We follow \cite{mo2019partnet} to utilize category mIoU and mAP ($50\%$ IoU threshold) as the semantic and instance segmentation metrics, respectively. We first calculate mIoU/mAP50 for each part category across all test shapes, and then average part mIoUs/mAP50s that belong to each object category to compute the object category mIoU/mAP50.

\vspace{-0.5em}
\subsection{Implementation Details}
\vspace{-0.5em}

For each 3D shape (i.e., ShapeNet~\cite{chang2015shapenet} mesh), we use BlenderProc~\cite{denninger2019blenderproc} to render 6 views of RGB-D images and segmentation masks with a resolution of $512\times512$. We unproject the images to the world space to obtain a fused point cloud with colors, normals, and ground truth part labels. The fused point clouds are used as the input for both our method and baseline approaches. 

For our method, we render each input point cloud into $K=10$ color images with Pytorch3D~\cite{ravi2020accelerating}. In few-shot experiments, we utilize 8 point clouds ($8\times10$ rendered images with 2D bounding boxes) of each object category for prompt tuning. The threshold $\tau$ in part instance grouping is empirically set to 0.3. 

\vspace{-0.5em}
\subsection{Comparison with Existing Methods}
\vspace{-0.5em}

\begin{table*}[t]
  \centering
  \scriptsize
  \setlength{\tabcolsep}{1.55pt}
  \caption{Semantic segmentation results on the PartNetE dataset. Object category mIoU($\%$) are shown. For 17 overlapping object categories, baseline models leverage additional 28k training shapes in the 45x8+28k setting. For the other 28 non-overlapping object categories, there are only 8 shapes per object category during training. Please refer to the supplementary for the full table of all 45 categories. }
    \begin{tabular}{x{0.075\linewidth}|c|cccccccc|c|cccccccc|c|c}
    \toprule
    \multirow{3}[4]{*}{\#3D data} & \multirow{3}[4]{*}{method} & \multicolumn{9}{c|}{Overlapping Categories}                           & \multicolumn{9}{c|}{Non-Overlapping Categories}                       &  \\
\cmidrule{3-20}          &       & \multirow{2}[2]{*}{Bottle} & \multirow{2}[2]{*}{Chair} & \multirow{2}[2]{*}{Display} & \multirow{2}[2]{*}{Door} & \multirow{2}[2]{*}{Knife} & \multirow{2}[2]{*}{Lamp} & Storage & \multirow{2}[2]{*}{Table} & Overall & \multirow{2}[2]{*}{Camera} & \multirow{2}[2]{*}{Cart} & Dis-  & \multirow{2}[2]{*}{Kettle} & Kitchen- & \multirow{2}[2]{*}{Oven} & Suit- & \multirow{2}[2]{*}{Toaster} & Overall & Overll \\
          &       &       &       &       &       &       &       & Furniture &       & (17)  &       &       & Penser &       & Pot   &       & case  &       & (28)  & (45) \\
    \midrule
    \multirow{3}[2]{*}{\makecell{few-shot w/ \\ extra data \\ (45x8+28k)}} & PointNet++~\cite{qi2017pointnet++} & 48.8  & 84.7  & 78.4  & 45.7  & 35.4  & 68.0  & 46.9  & \textbf{63.7}  & 55.6  & 6.5   & 6.4   & 12.1  & 20.9  & 15.8  & 34.3  & 40.6  & 14.7  & 25.4  & 36.8 \\
          & PointNext~\cite{qian2022pointnext} & 68.4  & \textbf{91.8}  & \textbf{89.4}  & 43.8  & 58.7  & 64.9  & \textbf{68.5}  & 52.1  & \textbf{58.5}  & 33.2  & 36.3  & 26.0  & 45.1  & 57.0  & 37.8  & 13.5  & 8.3   & \textbf{45.1}  & \textbf{50.2} \\
          & SoftGroup~\cite{vu2022softgroup} & 41.4  & 88.3  & 62.1  & \textbf{53.1}  & 31.3  & \textbf{82.2}  & 60.2  & 54.8  & 50.2  & 23.6  & 23.9  & 18.9  & 57.4  & 45.5  & 13.6  & 18.3  & 26.4  & 30.7  & 38.1  \\
    \midrule
    \multirow{6}[1]{*}{\makecell{few-shot \\ (45x8)}} & PointNet++~\cite{qi2017pointnet++} & 27.0  & 42.2  & 30.2  & 20.5  & 22.2  & 10.5  & 8.4   & 7.3   & 18.1  & 9.7   & 11.6  & 7.0   & 28.6  & 31.7  & 19.4  & 3.3   & 0.0   & 21.8  & 20.4 \\
          & PointNext~\cite{qian2022pointnext} & 67.6  & 65.1  & 53.7  & 46.3  & 59.7  & 55.4  & 20.6  & 22.1  & 39.2  & 26.0  & 47.7  & 22.6  & 60.5  & 66.0  & 36.8  & 14.5  & 0.0   & 41.5  & 40.6 \\
          & SoftGroup~\cite{vu2022softgroup} & 20.8  & 80.5  & 39.7  & 16.3  & 38.3  & 38.3  & 18.9  & 24.9  & 32.8  & 28.6  & 40.8  & 42.9  & 60.7  & 54.8  & 35.6  & 29.8  & 14.8  & 41.1  & 38.0 \\
        & ACD~\cite{gadelha2020label}   & 22.4  & 39.0  & 29.2  & 18.9  & 39.6  & 13.7  & 7.6   & 13.5  & 19.2  & 10.1  & 31.5  & 19.4  & 40.2  & 51.8  & 8.9   & 13.2  & 0.0   & 25.6  & 23.2 \\
          & Prototype~\cite{zhao2021few} & 60.1  & 70.8  & 67.3  & 33.4  & 50.4  & 38.2  & 30.2  & 25.7  & 41.1  & 32.0  & 36.8  & 53.4  & 62.7  & 63.3  & 36.5  & 35.5  & 10.1  & 46.3  & 44.3 \\
          & \textbf{Ours}  & \textbf{83.4}  & 85.3  & 84.8  & 40.8  & \textbf{65.2}  & 66.0  & 53.6  & 42.4  & \textbf{56.3}  & \textbf{58.3}  & \textbf{88.1}  & \textbf{73.7}  & \textbf{77.0}  & \textbf{69.6}  & \textbf{73.5}  & \textbf{70.4}  & \textbf{60.0}  & \textbf{61.3}  & \textbf{59.4} \\
          \midrule
      zero-shot & \textbf{Ours}  & 76.3  & 60.7  & 43.8  & 2.7   & 46.8  & 37.1  & 29.4  & 47.7  & 31.8  & 21.4  & 87.7  & 16.5  & 20.8  & 4.7   & 33.0  & 40.2  & 13.8  & 24.4  & 27.2 \\
    \bottomrule
    \end{tabular}%
    \label{table:semseg}
    \vspace{-1em}
\end{table*}%

\begin{table*}[t]
  \centering
  \scriptsize
  \setlength{\tabcolsep}{1.7pt}
  \caption{Instance segmentation results on the PartNetE dataset. Category mAP50 ($\%$) are shown. See supplementary for the full table.}
    \begin{tabular}{x{0.06\linewidth}|c|cccccccc|c|cccccccc|c|c}
    \toprule
    \multirow{3}[4]{*}{\#3D data} & \multirow{3}[4]{*}{method} & \multicolumn{9}{c|}{Overlapping Categories}                           & \multicolumn{9}{c|}{Non-Overlapping Categories}                       &  \\
\cmidrule{3-20}          &       & \multirow{2}[2]{*}{Bottle} & \multirow{2}[2]{*}{Chair} & \multirow{2}[2]{*}{Display} & \multirow{2}[2]{*}{Door} & \multirow{2}[2]{*}{Knife} & \multirow{2}[2]{*}{Lamp} & Storage & \multirow{2}[2]{*}{Table} & Overall & \multirow{2}[2]{*}{Camera} & \multirow{2}[2]{*}{Cart} & Dis-  & \multirow{2}[2]{*}{Kettle} & Kitchen- & \multirow{2}[2]{*}{Oven} & Suit- & \multirow{2}[2]{*}{Toaster} & Overall & Overll \\
          &       &       &       &       &       &       &       & Furniture &       & (17)  &       &       & Penser &       & Pot   &       & case  &       & (28)  & (45) \\
    \midrule
    \multirow{2}[1]{*}{45x8+28k} & PointGroup~\cite{jiang2020pointgroup} &  38.2  & 87.6  & 65.1  & \textbf{23.4}  & 19.3  & 62.7  & \textbf{49.1}  & \textbf{46.4}  & 41.7  & 8.6   & 29.2  & 24.0  & 61.3  & 59.4  & 13.8  & 15.6  & 7.0   & 24.6  & 31.0 \\
          & SoftGroup~\cite{vu2022softgroup} & 43.9  & \textbf{89.1}  & 68.7  & 21.2  & 27.2  & 63.3  & \textbf{49.1}  & 46.2  & \textbf{42.4}  & 0.7   & 28.4  & 26.4  & 63.8  & 59.3  & 16.4  & 13.5  & 7.5   & \textbf{25.6}  & \textbf{31.9}  \\ \midrule
    \multirow{3}[0]{*}{\makecell{few-shot \\ (45x8)}} & PointGroup~\cite{jiang2020pointgroup} & 8.0   & 77.2  & 16.7  & 3.7   & 15.6  & 9.8   & 0.0   & 0.0   & 14.6  & 4.7   & 28.5  & 30.7  & 52.1  & 57.0  & 0.0   & 0.0   & 0.0   & 16.8  & 16.0 \\
          & SoftGroup~\cite{vu2022softgroup} & 22.4  & 87.7  & 27.5  & 5.6   & 10.3  & 19.4  & 11.6  & 14.2  & 21.3  & 11.2  & 29.8  & 37.8  & 63.4  & 65.7  & 10.4  & 8.0   & 10.7  & 28.4  & 25.7   \\
          & \textbf{Ours}  & \textbf{79.4}  & 84.4  & \textbf{82.9}  & 17.9  & \textbf{43.9}  & \textbf{68.3}  & 32.8  & 32.3  & \textbf{42.5}  & \textbf{36.8}  & \textbf{83.3}  & \textbf{63.5}  & \textbf{75.4}  & \textbf{70.5}  & \textbf{64.5}  & \textbf{44.9}  & \textbf{38.4}  & \textbf{46.2}  & \textbf{44.8} \\ \midrule
    zero-shot     & \textbf{Ours}  & 75.5  & 54.5  & 32.9  & 1.3   & 22.1  & 35.8  & 10.9  & 36.6  & 20.9  & 8.4   & 79.3  & 9.3   & 18.3  & 1.1   & 25.9  & 34.2  & 4.5   & 16.2  & 18.0 \\
    \bottomrule
    \end{tabular}%
\label{table:inseg}
\vspace{-0.5em}
\end{table*}%

\subsubsection{Low-Shot Settings and Baseline Methods} \label{sec:settings_baselines}
\vspace{-0.5em}

 We consider three low-shot settings: (a) zero-shot: no 3D training/finetuning involved; (b) few-shot ($45\times8$): utilize only 8 shapes for each object category during training; (c) few-shot with additional data ($45\times8+28k$): utilize 28,367 shapes from PartNet~\cite{mo2019partnet} in addition to the $45\times8$ shapes during training. The 28k shapes cover 17 of the 45 object categories. Here, the last setting ($45\times8+28k$) describes a realistic setup, where we have large-scale part annotations for some common categories (17 categories in our case) but only a few shapes for the other categories. We aim to examine whether the 28k data of the 17 categories can help the part segmentation of the other 28 underrepresented categories. All settings are tested on the same test set. 

 We compare with PointNet++~\cite{qi2017pointnet++} and PointNext~\cite{qian2022pointnext} for semantic segmentation, and compare with PointGroup~\cite{jiang2020pointgroup} and SoftGroup~\cite{vu2022softgroup} for instance segmentation. We train four baseline approaches on the PartNetE dataset by taking point clouds with normals as input. For semantic segmentation, we follow \cite{mo2019partnet} to sample 10,000 points per shape as network input. For instance segmentation, we sample up to 50,000 points per shape. For each pair of baseline and setting, we train a single network. 

In addition to the four baselines mentioned above, we compare against two methods dedicated to few-shot 3D semantic segmentation: ACD~\cite{gadelha2020label} and Prototype~\cite{zhao2021few}. In ACD, we decompose the mesh of each 3D shape into approximate convex components with CoACD~\cite{wei2022approximate} and utilize the decomposition results for adding an auxiliary loss to the pipeline of PointNet++. In Prototype, we utilize the learned point features (by PointNext backbone) of few-shot shapes to construct 100 prototypes for each part category, which are then used to classify each point of test shapes. See supplementary for more details of baseline approaches.

\vspace{-1.5em}
\subsubsection{Evaluation Results}
\vspace{-0.5em}

\begin{figure}[t]
    \centering
    \vspace{-0.5em}
    \includegraphics[width=\linewidth]{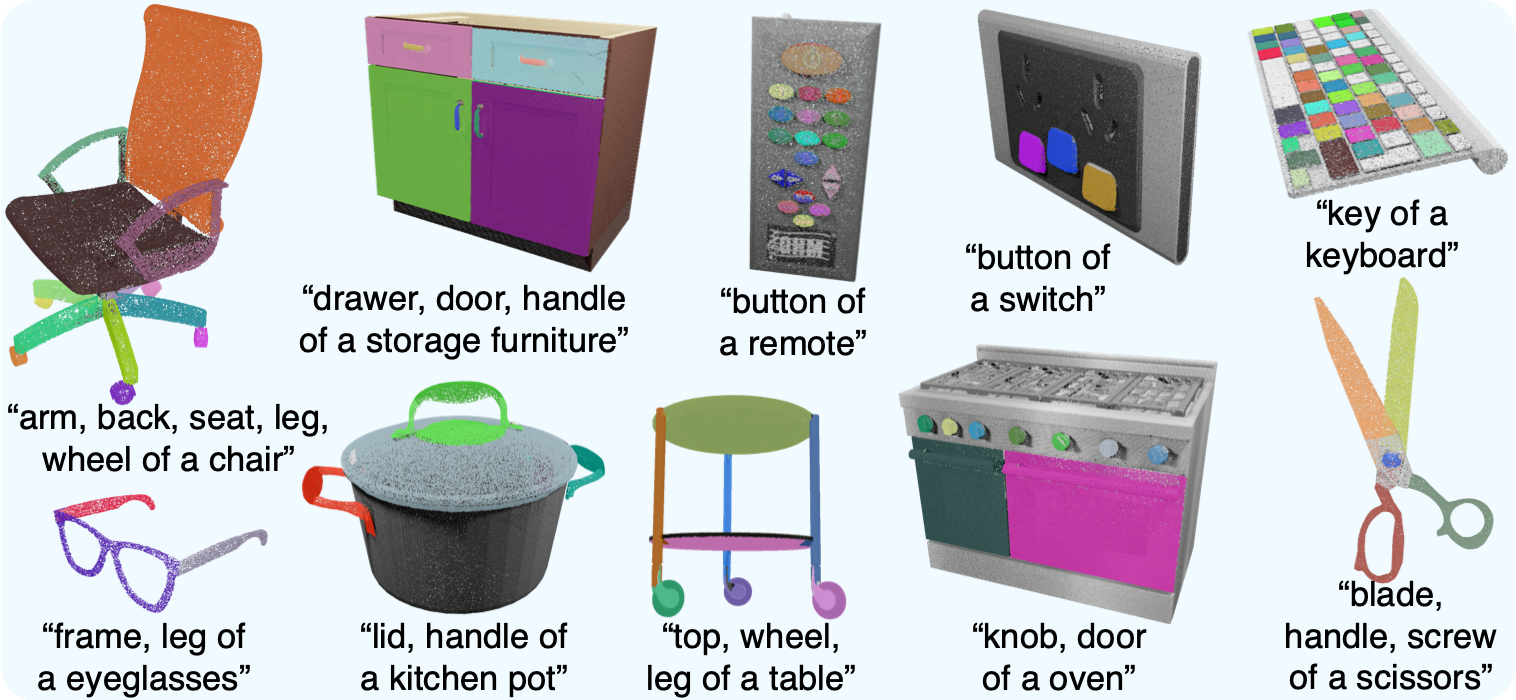}
    \caption{Instance segmentation results of our method (8-shot) on the PartNetE dataset. Different part instances are in different colors (zoom in for details).}
    \label{fig:instance_seg}
    \vspace{1.5em}
\end{figure}

Table~\ref{table:semseg} shows the results of semantic segmentation. Our method achieves impressive zero-shot performance on some common object categories (such as bottle, chair, and table), but also poor performances on certain categories (e.g., kettle). This is mainly due to the pretrained GLIP model may not understand the meaning of the text prompt (e.g., spout for kettles). After prompt tuning with 8-shot 3D data, our method achieves a 59.4\% mIoU and outperforms all baseline methods from the few-shot setting and even the $45\times8 + 28k$ setting. For the $45\times8+28k$ setting, baseline methods are trained with additional 28k shapes covering 17 categories. \textbf{For these overlapping categories, it's a fully-supervised setting, but our 8-shot version can achieve highly competitive overall mIoU (56.3\% vs.\ 58.5\%).} Note that the 28k training data is of limited help for the baselines to generalize to non-overlapping categories. Our method outperforms all baselines on non-overlapping categories by a large margin. The two few-shot strategies ACD and Prototype improve the performance of the original backbone, but there are still large gaps compared to our method. Please see Figure~\ref{fig:teaser} for example results of our methods and see supplementary for qualitative comparison.

Table~\ref{table:inseg} shows the results of instance segmentation. We observe similar phenomena as semantic segmentation. Our method achieves 18.0\% mAP50 for the zero-shot setting and 44.8\% mAP50 for the 8-shot setting, which outperforms all baseline approaches from both $45\times8$ and $45\times8+28k$ settings. See Figure~\ref{fig:instance_seg} for qualitative examples.

\subsection{Ablation Studies}

\vspace{-0.5em}
\paragraph{Proposed Components:}

\begin{table}[t]
  \centering
  \footnotesize
  \setlength{\tabcolsep}{4pt}
  \caption{Ablation study of the proposed components. We show the performances of both GLIP 2D detection (category mAP50) and 3D semantic segmentation (category mIoU) on three categories. *3D semantic segmentation is generated by assigning part labels to all visible points in bounding boxes. }
    \begin{tabular}{ccc|cc|cc|cc}
    \toprule
    BBox2    & Prompt & Feat  & \multicolumn{2}{c|}{Chair} & \multicolumn{2}{c|}{Kettle} & \multicolumn{2}{c}{Suitcase} \\
    3DSeg & Tuning & Aggre. & 2D    & 3D    & 2D    & 3D    & 2D    & 3D \\
    \midrule
          &       &       & 50.4  & 50.6* & 26.4  & 7.5*  & 31.9  & 21.1* \\
    \checkmark     &       &       & 50.4  & 60.7  & 26.4  & 20.8  & 31.9  & 40.2 \\
    \checkmark     & \checkmark     &       & 80.7  & 83.8  & 82.1  & 72.7  & 65.6  & 65.1 \\
    \checkmark     &       & \checkmark     & 52.3  & 64.5  & 32.2  & 25.9  & 36.4  & 49.1 \\
    \checkmark     & \checkmark     & \checkmark     & 82.4  & 85.3  & 84.3  & 77.0  & 68.9  & 70.4 \\
    \bottomrule
    \end{tabular}%
  \label{tab:ablation-component}%
  \vspace{2em}
\end{table}%

We ablate the proposed components, and the results are shown in Table~\ref{tab:ablation-component}. For the first row, we only utilize the pretrained GLIP model. In order to get 3D semantic segmentation, we assign part labels to all visible points within bounding boxes. The numbers indicate that this strategy is less effective than our proposed 3D voting and grouping module (second row). Moreover, without our proposed module, we are not able to get 3D instance segmentation. The second and third rows compare the impact of (8-shot) prompt tuning. We observe significant improvements, especially on the Kettle category, as the zero-shot GLIP model fails to understand the meaning of ``spout'' but it adapts to the definition after few-shot prompt tuning. The second and fourth rows compare our multi-view feature aggregation module. Without utilizing any extra data for finetuning, we leverage multi-view 3D priors to help the GLIP model better understand the input 3D shape and thus improve performance. After integrating all three modules, we achieve the final good performance (last row). 

\vspace{-1.5em}
\paragraph{Variations of Input Point Clouds:} Table~\ref{tab:ablation_input} evaluates the robustness of our method about variations of input point clouds. We observe that when the input point cloud is partial and does not cover all regions of the object, our method still performs well (second row). Also, we find that after removing the textures of the ShapeNet models and generating the input point cloud by using gray-scale images, our method can achieve good performance as well, suggesting that textures are less important in recognizing object parts. However, we find that the performance of our method may degrade when the input point cloud becomes sparse. On the one hand, sparse point clouds cause a larger domain gap for 2D renderings of point clouds. On the other hand, the sparsity makes it hard for our super point generation algorithm to produce good results. That being said, we want to point out that dense point clouds are already mostly available in our daily life (see Section~\ref{sec:real-world-demo}).

\begin{table}[t]
  \centering
  \footnotesize
  \caption{Ablation study of various input point clouds. We show the semantic segmentation results of the Chair category.}
    \begin{tabular}{c|ccc|c}
    \toprule
    setting & \# views & image reso. & texture & Chair mIoU ($\%$) \\
    \midrule
    original & 6     & $512\times512$ & w/    & 85.3 \\
    partial pc & 2     & $512\times512$ & w/    & 84.3 \\
    no texture & 6     & $512\times512$ & w/o   & 84.0 \\
    sparse pc & 6     & $128\times128$ & w/    & 82.4 \\
    sparse pc & 6     & $64\times64$ & w/    & 68.3 \\
    \bottomrule
    \end{tabular}
  \label{tab:ablation_input}
\end{table}

\begin{figure}[t]
    \centering
    \vspace{-0.5em}
    \includegraphics[width=\linewidth]{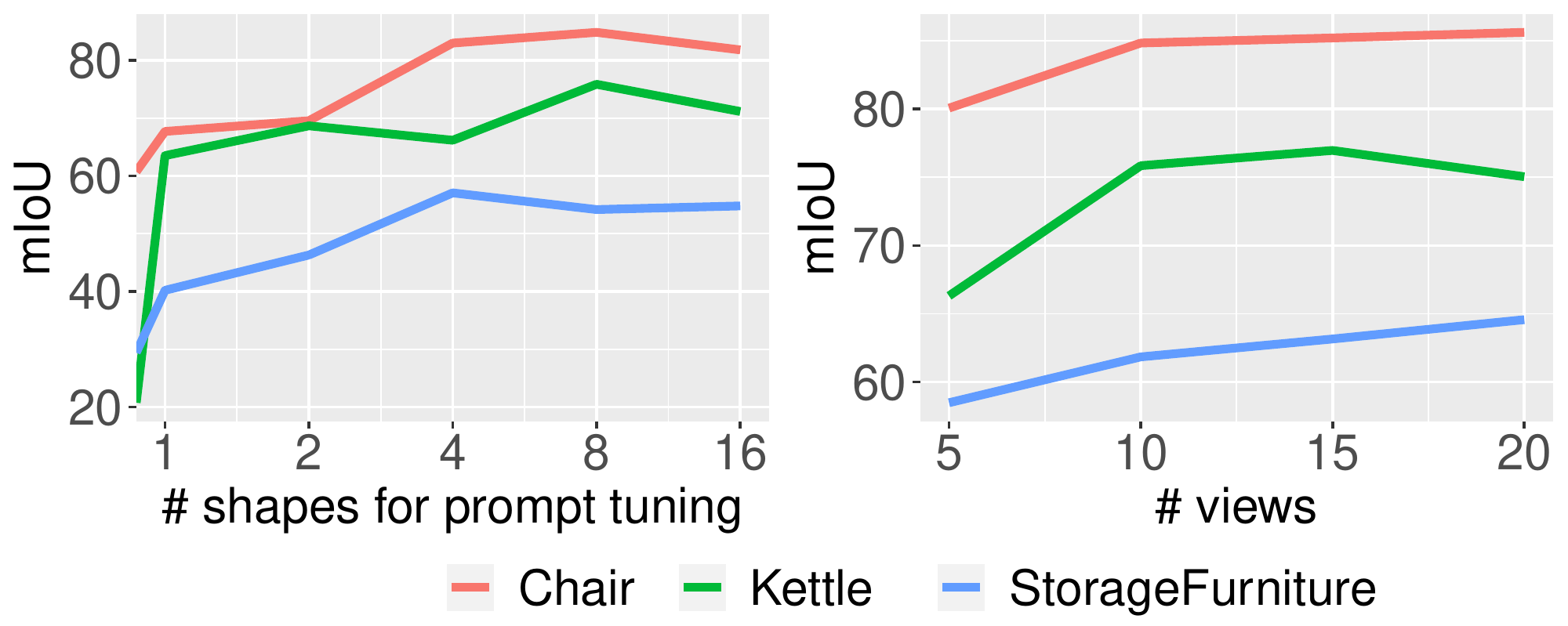}
    \caption{Ablation study of the number of shapes in prompt tuning and the number of 2D views ($K$). Category mIoU of 3D semantic segmentation on the PartNetE dataset are shown.}
    \label{fig:number_of_shapes_and_views}
    \vspace{1em}
\end{figure}

\vspace{-1.5em}
\paragraph{Number of Shapes in Prompt Tuning:} We ablate the number of shapes used for prompt tuning, and the results are shown in Figure~\ref{fig:number_of_shapes_and_views} (left). We observe that only using one single shape for prompt tuning can already improve the performance of the pretrained GLIP model a lot in some categories (e.g., Kettle). Also, after using more than 4 shapes, the gain from increasing the number of shapes slows down. We also find that prompt tuning is less effective for object categories that have richer appearance and structure variations (e.g., StorageFurniture). 

\vspace{-1.5em}
\paragraph{Number of 2D Views:} 
We render $K=10$ 2D views for each input point cloud in our main experiments. We ablate the value of $K$, and the results are shown in Figure~\ref{fig:number_of_shapes_and_views} (right). We observe a significant performance drop when $K$ is reduced to 5 and also a mild gain when using a larger $K$.

\vspace{-1.5em}
\paragraph{Early Fusion vs. Late Fusion:} 
In the last paragraph of Section~\ref{sec:multi_view_feature_aggregation}, we discuss two choices for multi-view feature aggregation: early fusion and late fusion. Table~\ref{tab:earlyfusion_vs_late_fusion} compares these two choices and verifies that late fusion will even degrade the performance while early fusion is helpful. 

\begin{table}[t]
  \centering
   \footnotesize
   \vspace{-2em}
  \caption{Early vs.\ late fusion in multi-view feature aggregation. We compare GLIP detection (mAP50) on the Suitcase category.}
    \begin{tabular}{ccc}
    \toprule
    w/o fusion & early fusion & late fusion \\
    \midrule
     65.6     &   68.9    &  47.3 \\
    \bottomrule
    \end{tabular}%
  \label{tab:earlyfusion_vs_late_fusion}%
\end{table}%

\vspace{-1.5em}
\paragraph{GLIP vs.\ CLIP:}
We have also considered using other pretrained vision-language models, such as CLIP~\cite{radford2021learning}. However, we find that the pretrained CLIP model fails to recognize fine-grained object parts and has difficulty generating region-level output. See supplementary for details.

\vspace{-0.3em}
\subsection{Real-World Demo}
\vspace{-0.3em}
\label{sec:real-world-demo}
\begin{figure}[t]
    \vspace{-0.3em}
    \centering
    \includegraphics[width=\linewidth]{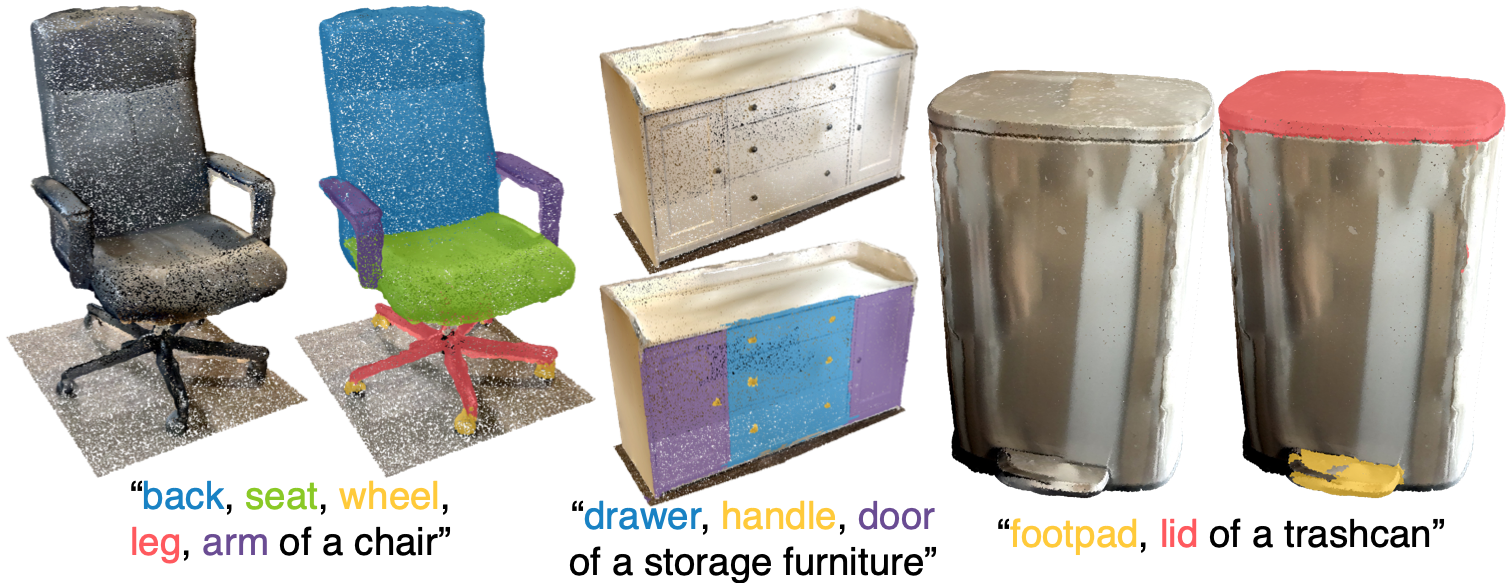}
    \caption{Each pair shows a captured point cloud by iPhone (left) and the semantic segmentation result of our method (right).}
    \label{fig:real_world_demo}
    \vspace{1em}
\end{figure}

Thanks to the strong generalizability of the GLIP model, our method can be directly deployed in the real world without a significant domain gap. As shown in Figure~\ref{fig:real_world_demo}, we use an iPhone 12 Pro Max, equipped with a LiDAR sensor, to capture a video and feed the fused point cloud to our method. We observe similar performances as in our synthetic experiments. Please note that existing 3D networks are sensitive to the input format. For example, they assume objects are normalized in per-category canonical poses. Also, they need to overcome the significant domain gap, making it hard to deploy them directly in real scenarios. See supplementary for more details.

\vspace{-0.3em}
\section{Discussion and Limitations}
\vspace{-0.3em}
\label{sec:conclusion}

The current pipeline utilizes predicted bounding boxes from the GLIP model. We notice that GLIPv2~\cite{zhang2022glipv2} has 2D segmentation capabilities, but their pretrained model is not released at the time of submission. We admit that it will be more natural to use 2D segmentation results, which are more accurate than bounding boxes, from pretrained models. However, we want to point out that it is still non-trivial to get 3D instance segmentation even from multi-view 2D segmentation, and all components of our proposed method would still be useful (with necessary adaptations). A bigger concern is that our method cannot handle the interior points of objects. It also suffers from long running time due to point cloud rendering and multiple inferences of the GLIP model. Therefore, using our method to distill the knowledge of 2D VL models and train 3D foundation models is a promising future direction, which may lead to more efficient inferences.

\renewcommand{\thesection}{S}
\renewcommand{\thetable}{S\arabic{table}}
\renewcommand{\thefigure}{S\arabic{figure}}
\renewcommand{\theequation}{S.\arabic{equation}}

\begin{table*}[t]
  \scriptsize
  \setlength{\tabcolsep}{1.7pt}
  \centering
  \caption{The table shows the statistics of the PartNetE dataset: category name, part names, number of few-shot shapes, test shapes, and additional training shapes (if applicable). The 17 overlapping object categories are bolded.}
    \begin{tabular}{llrrr||llrrr}
    \toprule
    \multicolumn{1}{c}{category} & \multicolumn{1}{c}{parts} & \multicolumn{1}{c}{few-shot} & \multicolumn{1}{c}{test} & \multicolumn{1}{c||}{extra-train} & \multicolumn{1}{c}{category} & \multicolumn{1}{c}{parts} & \multicolumn{1}{c}{few-shot} & \multicolumn{1}{c}{test} & \multicolumn{1}{c}{extra-train} \\
    \midrule
    \textbf{Bottle} & lid   & 8     & 49    & 471   & \textbf{Microwave} & display, door, handle, button & 8     & 8     & 234 \\
    Box   & lid   & 8     & 20    & 0     & Mouse & button, cord, wheel & 8     & 6     & 0 \\
    Bucket & handle & 8     & 28    & 0     & Oven  & door, knob & 8     & 22    & 0 \\
    Camera & button, lens & 8     & 29    & 0     & Pen   & cap, button & 8     & 40    & 0 \\
    Cart  & wheel & 8     & 53    & 0     & Phone & lid, button & 8     & 10    & 0 \\
    \textbf{Chair} & arm, back, leg, seat, wheel & 8     & 73    & 8000  & Pliers & leg   & 8     & 17    & 0 \\
    \textbf{Clock} & hand  & 8     & 23    & 593   & Printer & button & 8     & 21    & 0 \\
    CoffeeMachine & button, container, knob, lid & 8     & 46    & 0     & \textbf{Refrigerator} & door, handle & 8     & 36    & 195 \\
    \textbf{Dishwasher} & door, handle & 8     & 40    & 179   & Remote & button & 8     & 41    & 0 \\
    Dispenser & head, lid & 8     & 49    & 0     & Safe  & door, switch, button & 8     & 22    & 0 \\
    \textbf{Display} & base, screen, support & 8     & 29    & 954   & \textbf{Scissors} & blade, handle, screw & 8     & 39    & 60 \\
    \textbf{Door} & frame, door, handle & 8     & 28    & 237   & Stapler & body, lid & 8     & 15    & 0 \\
    Eyeglasses & body, leg & 8     & 57    & 0     & \textbf{StorageFurniture} & door, drawer, handle & 8     & 338   & 2260 \\
    \textbf{Faucet} & spout, switch & 8     & 76    & 681   & Suitcase & handle, wheel & 8     & 16    & 0 \\
    FoldingChair & seat  & 8     & 18    & 0     & Switch & switch & 8     & 62    & 0 \\
    Globe & sphere & 8     & 53    & 0     & \textbf{Table} & door, drawer, leg, tabletop, wheel, handle & 8     & 93    & 9799 \\
    Kettle & lid, handle, spout & 8     & 21    & 0     & Toaster & button, slider & 8     & 17    & 0 \\
    \textbf{Keyboard} & cord, key & 8     & 29    & 165   & Toilet & lid, seat, button & 8     & 61    & 0 \\
    KitchenPot & lid, handle & 8     & 17    & 0     & \textbf{TrashCan} & footpedal, lid, door & 8     & 62    & 358 \\
    \textbf{Knife} & blade & 8     & 36    & 505   & USB   & cap, rotation & 8     & 43    & 0 \\
    \textbf{Lamp} & base, body, bulb, shade & 8     & 37    & 3246  & WashingMachine & door, button & 8     & 9     & 0 \\
    \textbf{Laptop} & keyboard, screen, shaft, touchpad, camera & 8     & 47    & 430   & Window & window & 8     & 50    & 0 \\
    Lighter & lid, wheel, button & 8     & 20    & 0     & \textbf{45 in total} & \textbf{103 in total} & \textbf{360} & \textbf{1,906} & \textbf{28,367} \\
    \bottomrule
    \end{tabular}%
  \label{tab:dataset_stat}%
  \vspace{-1em}
\end{table*}%

\section{Supplementary Material}

In this supplementary material, we first present more details of the proposed dataset, PartNet-Ensembled (Sec. ~\ref{sec:PartNetE}). We then show more results of real-world demos (Sec.~\ref{sec:supp-real-world-demo}) and visualization of various ablation studies (Sec.~\ref{sec:vis-ablation}). We also compare CLIP and GLIP on object part recognition (Sec.~\ref{sec:clip}) and show qualitative comparisons between our method and baseline approaches (Sec.~\ref{sec:qualitative}). Finally, we present implementation details of baseline approaches (Sec.~\ref{sec:baselines}) and full tables of quantitative comparisons (Sec.~\ref{sec:full_table}).

\subsection{PartNet-Ensembled Dataset}
\label{sec:PartNetE}

Table~\ref{tab:dataset_stat} shows the statistics of the proposed PartNet-Ensembled (PartNetE) dataset. The few-shot and test shapes come from PartNet-Mobility~\cite{xiang2020sapien}, and the additional training shapes come from PartNet~\cite{mo2019partnet}. All three sets share consistent part definitions. To construct a diverse, clear, and consistent 3D object-part dataset, we select a subset of 100 object parts from the original PartNet and PartNet-Mobility annotations, and manually annotate three additional parts (i.e., Kettle spout, KitchenPot handle, and Mouse cord). Specifically, we filter out extremely fine-grained parts (e.g., ``back\_frame\_vertical\_bar" for chairs), ambiguous parts, inconsistently annotated parts, and rarely seen parts of the original datasets. As a result, each object category contains 1-6 parts in our PartNetE dataset, covering both common coarse-grained parts (e.g., chair back and tabletop) and fine-grained parts (e.g., wheel, handle, button, knob, switch, touchpad) that may be useful in downstream tasks such as robotic manipulation. For shapes from PartNet-Mobility, they have textures, while for shapes from PartNet, they do not. The unbalanced data distribution is a critical issue when using the additional 28k training shapes. We may have nearly 10k shapes for common categories (e.g., Table) but only 8 for some non-overlapping categories. We believe our dataset could benefit future works on low-shot and text-driven 3D part annotation, which do not rely on large-scale supervised learning to infer part definitions. 

\subsection{Real-World Demo}
\label{sec:supp-real-world-demo}
\begin{figure*}[t]
    \centering
    \includegraphics[width=\linewidth]{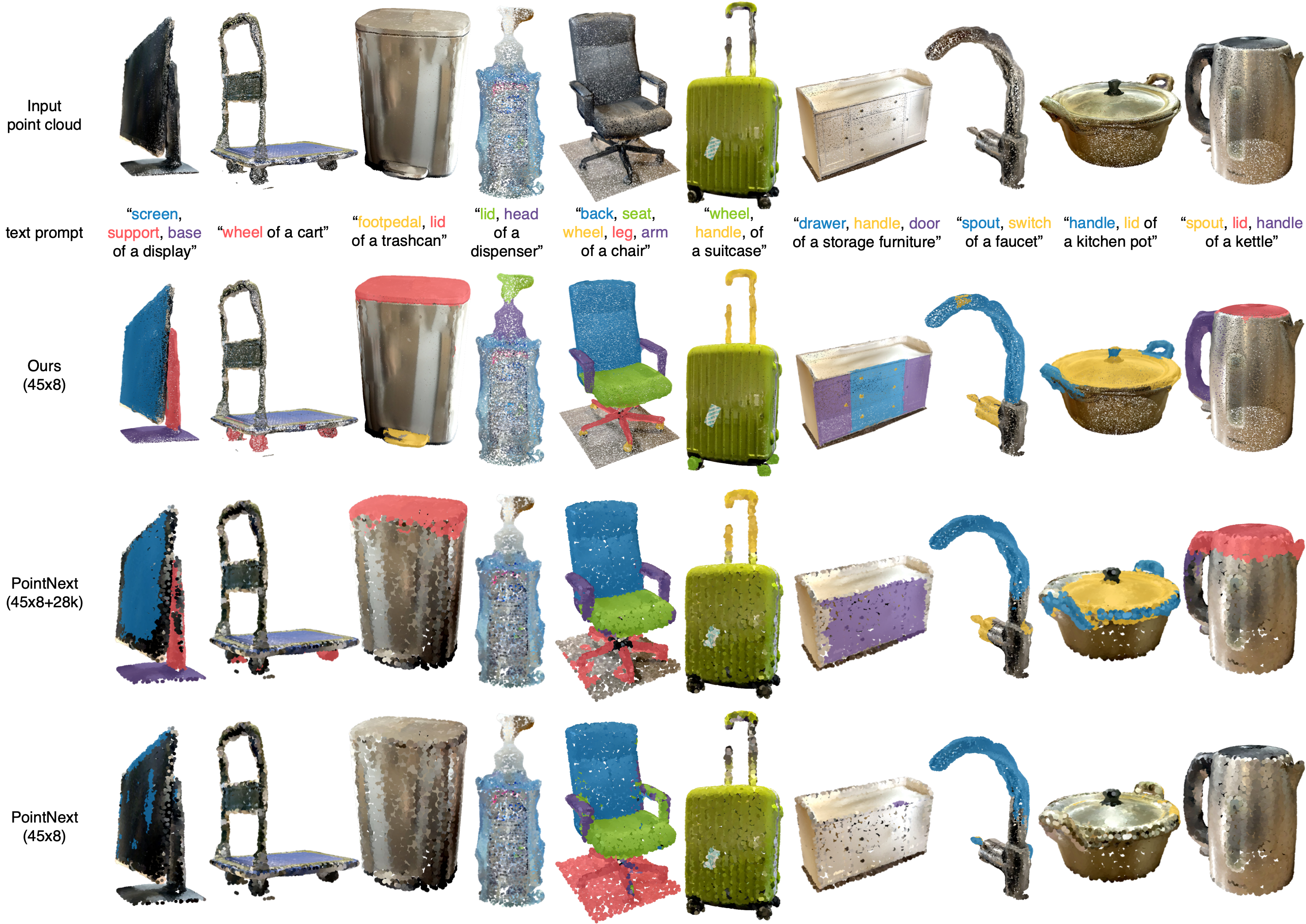}
    \caption{Real-world demo: iPhone-scanned point clouds (first row), text prompt for our method (second row), results of our method and baseline approaches (third to fifth rows). ``45x8'' indicates the few-shot setting, where the model is trained with 8 shapes per object category. ``45x8+28k'' indicates the setting where the additional 28k shapes are used for training. Zoom in for details.}
    \label{fig:sup_real_world}
\end{figure*}

Figure~\ref{fig:sup_real_world} shows more examples when our method and baseline approaches are applied to point clouds captured by an iPhone 12 Pro Max equipped with a LiDAR sensor. Specifically, we utilize the APP ``polycam" to scan daily objects and generate fused point clouds with color. We use MeshLab to remove ground points and compute point normals. For baseline approaches, we randomly sample 10,000 points as input.

As shown in the figure, our method can directly generalize to iPhone-scanned point clouds without significant domain gaps, while baseline methods perform poorly. For PointNext~\cite{qian2022pointnext} of the ``45x8+28k'' setting (third row), it uses the additional 28k training data but still fails to recognize many parts (e.g., cart wheels, trashcan footpedal, lid and head of the dispenser, chair wheels, suitcase wheels, drawers and handles of the storage furniture, handle of the kettle). The few-shot version (fourth row) performs even worse and can only identify a few parts.

\subsection{Visualization of Ablation Studies}

\begin{figure*}[t]
    \centering
    \includegraphics[width=\linewidth]{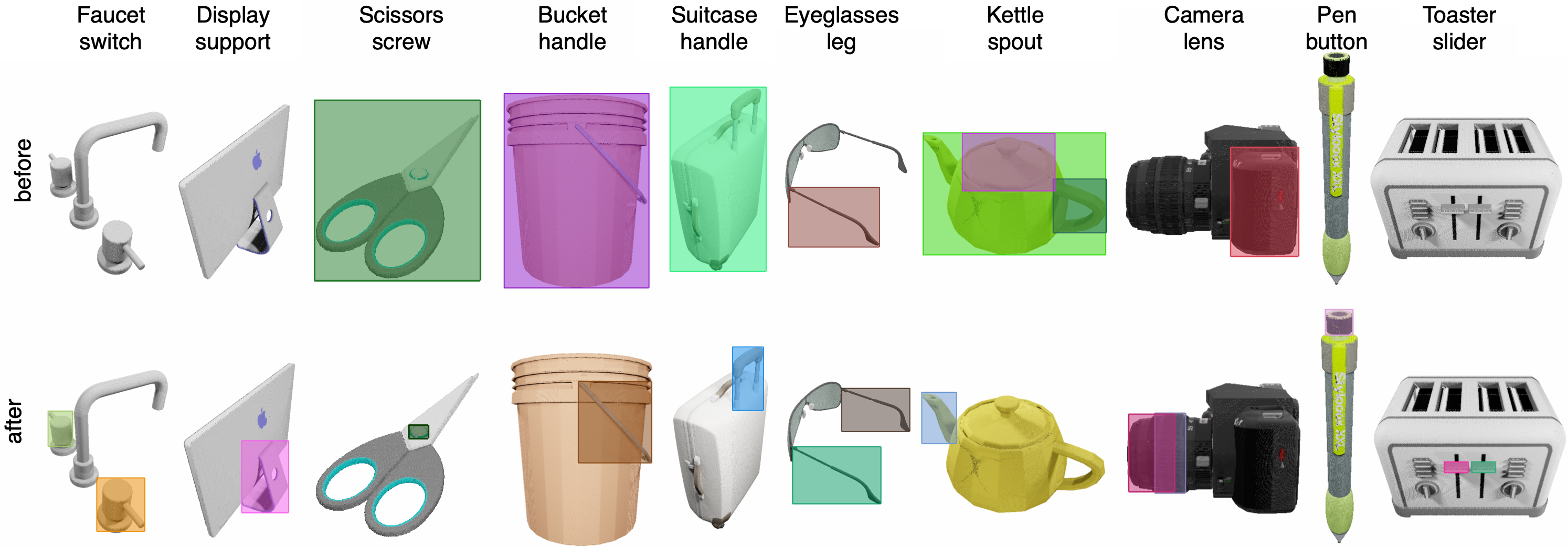}
    \caption{Ablation study of few-shot prompt tuning. First row: 2D part detection results of the GLIP pretrained model (zero-shot). Second row: detection results after 8-shot prompt tuning.}
    \label{fig:supp-prompt-tuning}
    \vspace{-1em}
\end{figure*}

\paragraph{Few-Shot Prompt Tuning} Figure~\ref{fig:supp-prompt-tuning} shows the comparison before and after few-shot prompt tuning. The pretrained GLIP model (first row) fails to understand the meaning of many part names. However, after prompt tuning with only one or a few segmented 3D shapes (second row), the GLIP model quickly adapts to part definitions and can generalize to unseen instances.

\label{sec:vis-ablation}

\begin{figure*}[t]
    \centering
    \includegraphics[width=\linewidth]{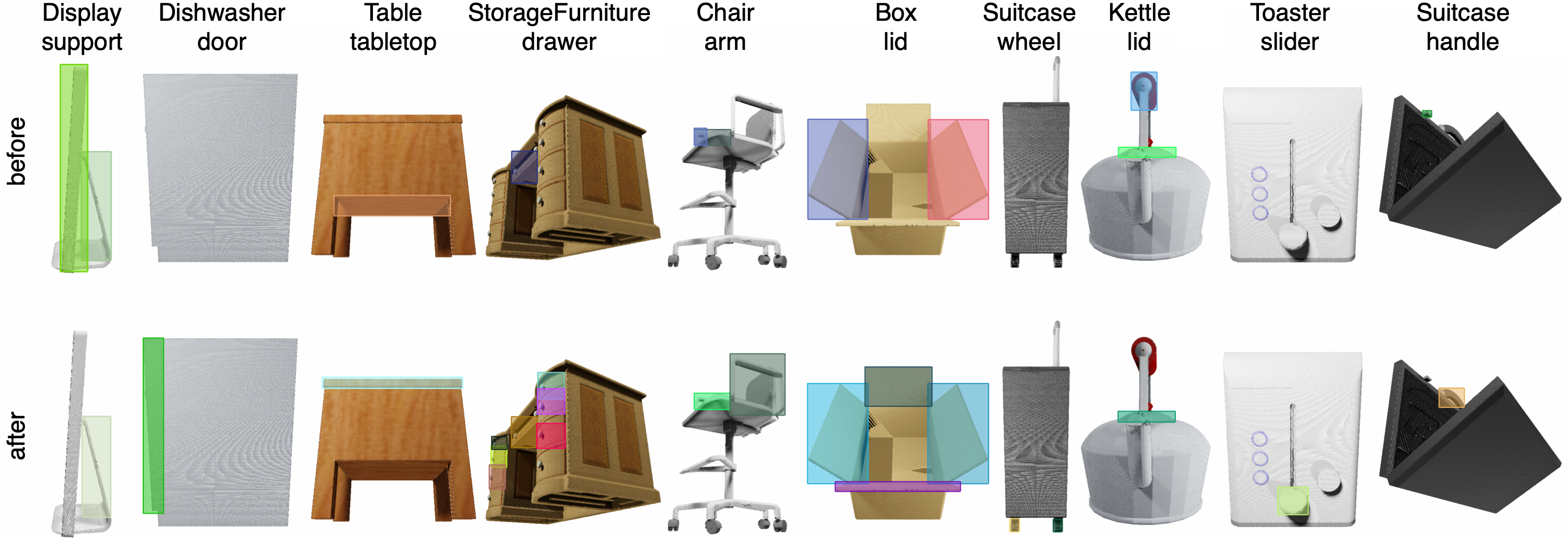}
    \caption{Ablation study of multi-view visual feature aggregation. First row: 2D part detection results without the multi-view visual feature aggregation. Second row: detection results with the multi-view feature aggregation. Both models are prompt-tuned.}
    \label{fig:sup-multi-view-aggregation}
    \vspace{-1em}
\end{figure*}

\paragraph{Multi-View Visual Feature Aggregation}
Figure~\ref{fig:sup-multi-view-aggregation} shows the comparison with and without multi-view visual feature aggregation. When there is no multi-view visual feature aggregation (first row), the GLIP model fails to detect parts from some unfamiliar camera views. However, after aggregating visual features from multiple views (second row), the GLIP model can comprehensively understand input 3D shapes and make more accurate predictions for those unfamiliar views.

\begin{figure*}[t]
    \centering
    \includegraphics[width=\linewidth]{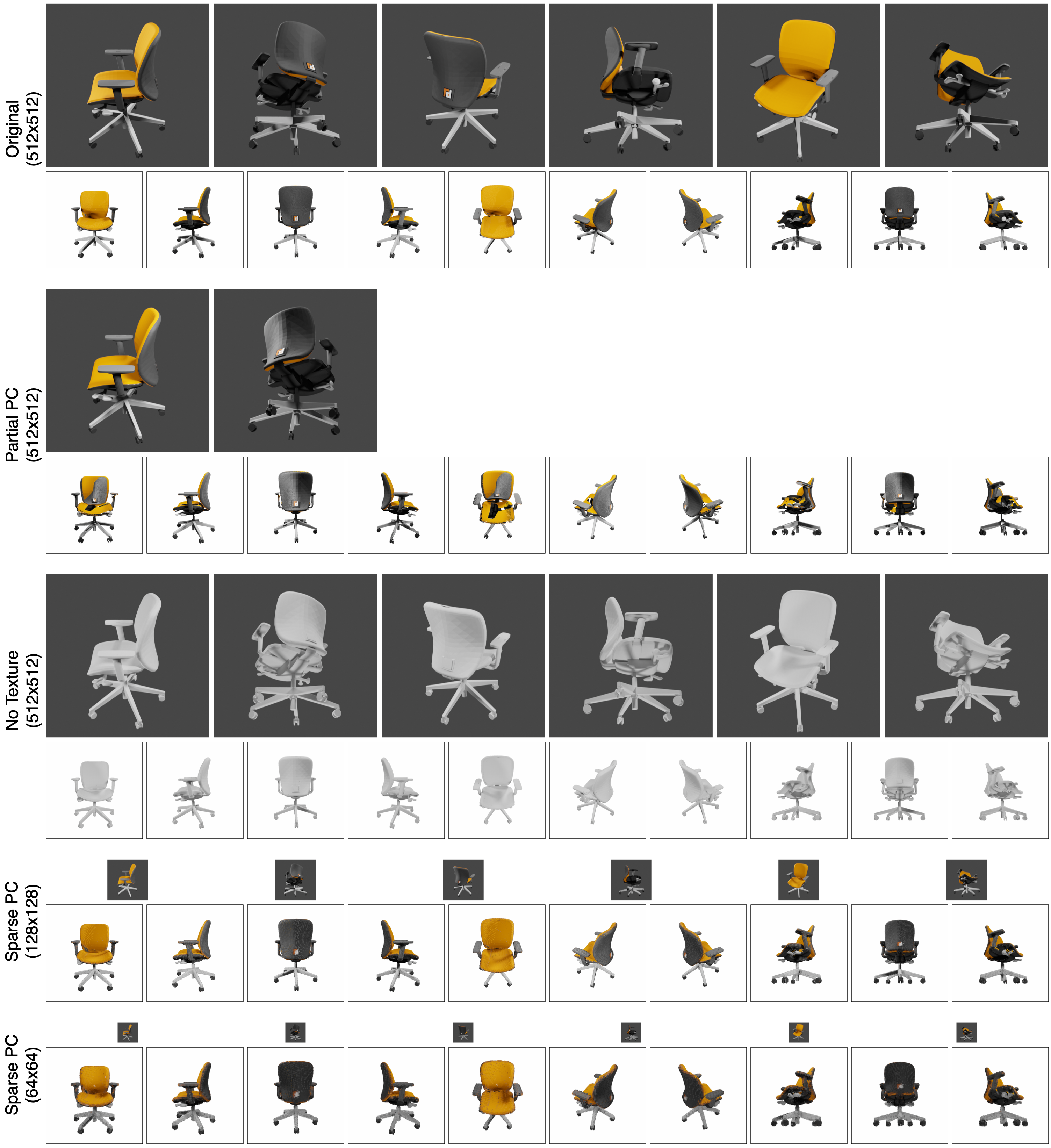}
    \caption{Five variants of input point clouds. For each variant, the first row shows mesh renderings by BlenderProc~\cite{denninger2019blenderproc}, which are used to fuse and generate the input point cloud. The resolutions of the images are shown in parentheses. The second row shows renderings of the input point cloud by Pytorch3D~\cite{ravi2020accelerating}, which are fed to the GLIP model. The image resolution is 800x800. Artifacts of point cloud renderings (last row) can be seen when zoomed in.}
    \vspace{-1em}
    \label{fig:supp-input-pc}
\end{figure*}

\vspace{-1em}
\paragraph{Variations of Input Point Clouds} To evaluate the robustness of our method, we have tried multiple variations of input point clouds (see Table~\ref{tab:ablation_input}). Figure~\ref{fig:supp-input-pc} exemplifies 2D images used to generate input point clouds and point cloud renderings fed to the FLIP model. In the original setting, we use 6 RGB-D images with a resolution of 512x512 to generate the fused point cloud, which is then projected to 10 2D images with a resolution of 800x800. Note that when point clouds are sparse, we increase the point size to reduce the artifacts of point cloud renderings. Please zoom in to find the differences between point cloud renderings. As shown in Table~\ref{tab:ablation_input}, our proposed method is robust against various input point cloud variations.

\subsection{CLIP vs.\ GLIP}
\label{sec:clip}

\begin{figure*}[t]
    \centering
    \includegraphics[width=\linewidth]{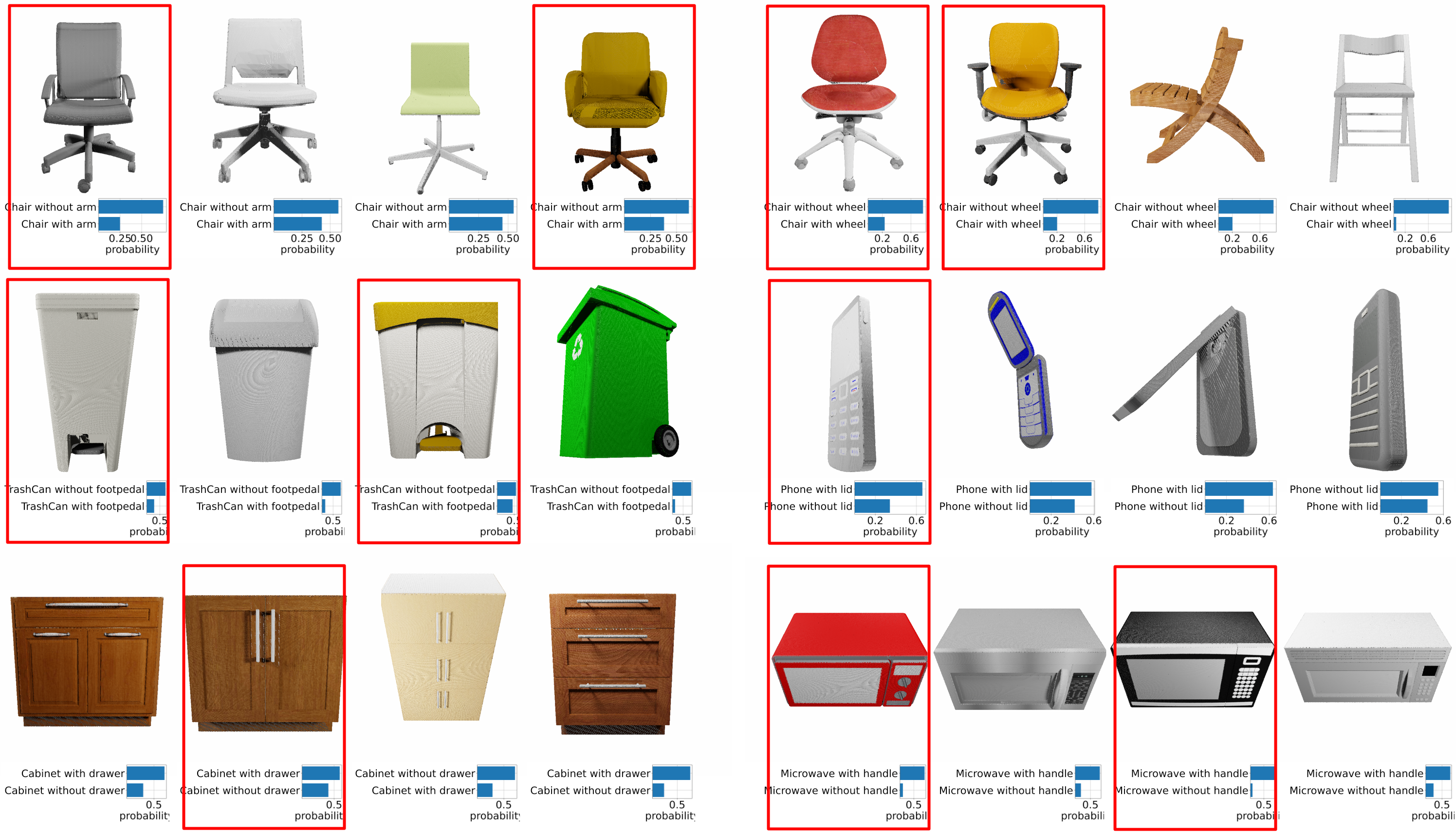}
    \caption{We perform binary classification using CLIP~\cite{radford2021learning}. CLIP fails to identify whether an object has a part. Incorrect predictions are highlighted with red rectangles.}
    \label{fig:supp-CLIP}
    \vspace{-1em}
\end{figure*}
We have also considered using other pretrained vision-language models, such as CLIP~\cite{radford2021learning}, to help with part segmentation tasks. However, the CLIP model mainly focuses on the image classification task and cannot directly generate region-level output (e.g., 2D segmentation masks or bounding boxes). Moreover, as shown in Figure~\ref{fig:supp-CLIP}, we find that the pretrained CLIP model fails to tell whether an object has a fine-grained part. We conjecture that the CLIP model is pretrained using image-level supervision, with fewer supervision signals about object parts. In contrast, the GLIP model is pretrained on 2D detection and grounding tasks and is thus more sensitive to fine-grained object parts. As a result, the GLIP model is more suitable for our 3D part segmentation task.

\subsection{Qualitative Comparison on PartNetE}
\label{sec:qualitative}
\begin{figure*}[t]
    \centering
    \includegraphics[width=\linewidth]{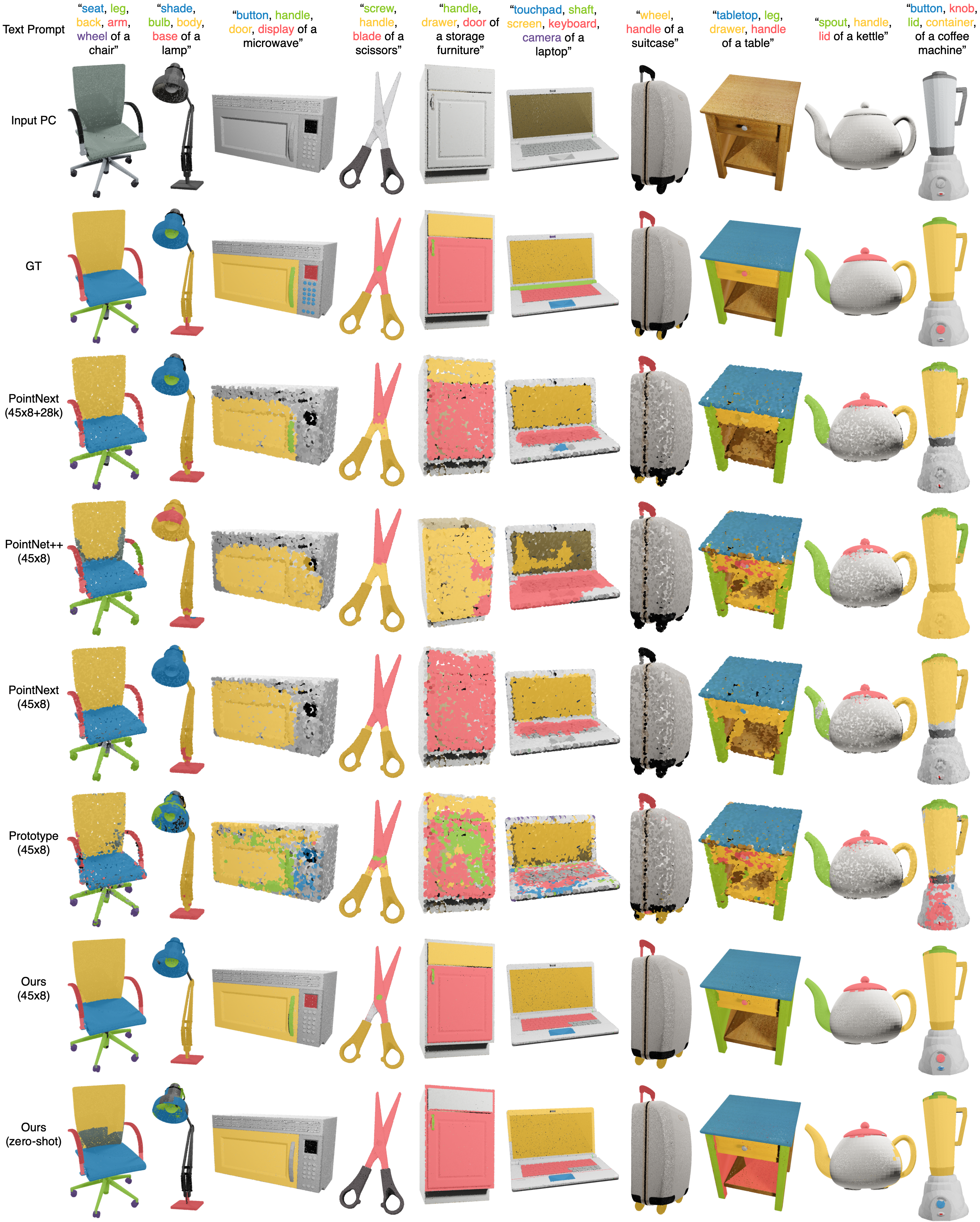}
    \caption{Qualitative comparison between our method and baseline approaches on the PartNetE dataset. Semantic segmentation results are shown. For baseline approaches, we randomly sample 10,000 points as input. ``45x8'' indicates the few-shot setting, where the model is trained with 8 shapes per object category. ``45x8+28k'' indicates the setting where the additional 28k shapes are used for training .}
\label{fig:sup-qualitative}
\vspace{-4em}
\end{figure*}

Figure~\ref{fig:sup-qualitative} shows the qualitative comparison between our method and baseline approaches. Our few-shot version (45x8) outperforms all existing few-shot methods and even produces better results than the ``45x8+28k'' version of PointNext, where the additional 28k 3D shapes are used for training. In particular, our method is good at detecting small object parts (i.e., wheel, bulb, screw, handle, knob, and button). Without any 3D training, our zero-shot version also achieves impressive results.

\begin{table*}[t]
  \centering
  \scriptsize
  \setlength{\tabcolsep}{2.5pt}
  \setlength\extrarowheight{1pt}
  \caption{Full table (1/2) of semantic segmentation results on the PartNetE dataset. Category mIoUs are shown. For 17 overlapping object categories, baseline models leverage additional 28k training shapes in the 45x8+28k setting. For the other 28 non-overlapping object categories, there are only 8 shapes per object category during training.}
    \begin{tabular}{c|cc|rrr|rrrrrr|r}
    \toprule
    \multirow{53}[0]{*}{\begin{sideways}Overlapping Categories (17)\end{sideways}} & \multicolumn{2}{c|}{} & \multicolumn{3}{c|}{few-shot w/ additional data (45x8+28k)} & \multicolumn{6}{c|}{few-shot (45x8)} & \multicolumn{1}{c}{zero-shot} \\
\cmidrule{2-13}          & category & part  & \multicolumn{1}{c}{PointNet++~\cite{qi2017pointnet++}} & \multicolumn{1}{c}{PointNext~\cite{qian2022pointnext}} & \multicolumn{1}{c|}{SoftGroup~\cite{vu2022softgroup}} & \multicolumn{1}{c}{PointNet++~\cite{qi2017pointnet++}} & \multicolumn{1}{c}{PointNext~\cite{qian2022pointnext}} & \multicolumn{1}{c}{SoftGroup~\cite{vu2022softgroup}} & \multicolumn{1}{c}{ACD~\cite{gadelha2020label}} & \multicolumn{1}{c}{Prototype~\cite{zhao2021few}} & \multicolumn{1}{c|}{Ours} & \multicolumn{1}{c}{Ours} \\
\cmidrule{2-13}          & Bottle & lid   & 48.8  & 68.4  & 41.4  & 27.0  & 67.6  & 20.8  & 22.4  & 60.1  & \textbf{83.4} & 76.3 \\
\cline{2-13}          & \multirow{5}[0]{*}{Chair} & arm   & 83.5  & 88.6  & \textbf{89.7} & 29.5  & 68.6  & 67.8  & 27.6  & 58.7  & 74.1  & 34.6 \\
          &       & back  & 89.0  & \textbf{93.4} & 92.2  & 59.7  & 89.5  & 86.5  & 60.6  & 83.7  & 89.7  & 25.3 \\
          &       & leg   & 85.5  & \textbf{94.0} & 83.5  & 51.7  & 70.0  & 84.9  & 42.8  & 73.0  & 89.0  & 76.3 \\
          &       & seat  & 85.7  & \textbf{90.5} & 81.8  & 61.0  & 80.8  & 76.6  & 53.4  & 70.9  & 81.4  & 75.3 \\
          &       & wheel & 79.7  & 92.6  & \textbf{94.4} & 9.0   & 16.7  & 86.6  & 10.7  & 67.9  & 92.6  & 92.2 \\
\cline{2-13}          & Clock & hand  & 19.2  & 28.4  & 2.5   & 0.0   & 0.0   & 6.0   & 0.0   & 10.5  & \textbf{37.6} & 26.7 \\
\cline{2-13}          & \multirow{2}[0]{*}{Dishwasher} & door  & 59.3  & \textbf{81.5} & 50.7  & 55.6  & 73.9  & 54.2  & 50.6  & 68.6  & 71.2  & 20.5 \\
          &       & handle & 39.6  & \textbf{56.8} & 55.3  & 0.0   & 0.0   & 30.1  & 0.0   & 28.0  & 53.8  & 0.0 \\
\cline{2-13}          & \multirow{3}[0]{*}{Display} & base  & 88.1  & \textbf{97.1} & 94.5  & 48.9  & 82.3  & 50.5  & 36.9  & 76.9  & 97.0  & 70.1 \\
          &       & screen & 80.4  & \textbf{87.6} & 49.6  & 40.1  & 78.8  & 46.1  & 42.1  & 73.6  & 73.9  & 61.2 \\
          &       & support & 66.5  & \textbf{83.4} & 42.3  & 1.5   & 0.0   & 22.6  & 8.4   & 51.5  & 83.4  & 0.0 \\
\cline{2-13}          & \multirow{3}[0]{*}{Door} & frame & 48.2  & 50.0  & 42.6  & 22.6  & \textbf{65.6} & 23.4  & 23.5  & 49.1  & 20.9  & 1.0 \\
          &       & door  & 60.2  & \textbf{75.7} & 65.7  & 38.9  & 73.3  & 16.6  & 33.1  & 50.1  & 70.8  & 7.1 \\
          &       & handle & 28.6  & 5.7   & \textbf{51.0} & 0.0   & 0.0   & 8.9   & 0.0   & 1.2   & 30.7  & 0.0 \\
\cline{2-13}          & \multirow{2}[0]{*}{Faucet} & spout & 80.1  & \textbf{90.4} & 82.6  & 31.2  & 67.2  & 50.4  & 31.4  & 62.1  & 79.0  & 12.7 \\
          &       & switch & 54.3  & \textbf{79.5} & 54.1  & 10.8  & 33.3  & 18.5  & 16.9  & 29.9  & 63.8  & 0.9 \\
\cline{2-13}          & \multirow{2}[0]{*}{Keyboard} & cord  & 82.3  & 6.1   & 78.0  & 0.0   & 0.0   & 57.1  & 0.0   & 31.2  & \textbf{83.9} & 74.6 \\
          &       & key   & 66.7  & \textbf{83.8} & 39.8  & 31.5  & 69.2  & 50.2  & 52.2  & 58.5  & 23.3  & 0.0 \\
\cline{2-13}          & Knife & blade & 35.4  & 58.7  & 31.3  & 22.2  & 59.7  & 38.3  & 39.6  & 50.4  & \textbf{65.2} & 46.8 \\
\cline{2-13}          & \multirow{4}[0]{*}{Lamp} & base  & 77.5  & 72.8  & \textbf{92.8} & 20.5  & 82.0  & 48.7  & 6.0   & 56.2  & 90.3  & 84.5 \\
          &       & body  & 64.5  & 65.8  & 78.2  & 17.5  & 64.4  & 40.5  & 27.3  & 59.0  & \textbf{79.2} & 0.0 \\
          &       & bulb  & 51.4  & 35.2  & \textbf{66.3} & 0.0   & 0.0   & 12.2  & 0.0   & 4.4   & 10.2  & 12.6 \\
          &       & shade & 78.5  & 85.7  & \textbf{91.5} & 4.1   & 75.1  & 52.0  & 21.5  & 33.1  & 84.5  & 51.3 \\
\cline{2-13}          & \multirow{5}[0]{*}{Laptop} & keyboard & 66.4  & \textbf{70.4} & 25.1  & 22.0  & 40.6  & 41.9  & 20.0  & 48.3  & 60.1  & 48.0 \\
          &       & screen & 79.0  & \textbf{83.0} & 33.9  & 28.4  & 79.9  & 42.6  & 35.5  & 68.2  & 62.8  & 71.2 \\
          &       & shaft & \textbf{27.7} & 0.0   & 19.6  & 0.0   & 0.0   & 13.4  & 0.0   & 8.7   & 3.0   & 0.0 \\
          &       & touchpad & \textbf{27.3} & 9.1   & 9.4   & 0.0   & 0.0   & 7.8   & 0.0   & 13.6  & 20.6  & 11.4 \\
          &       & camera & \textbf{76.6} & 0.0   & 4.1   & 0.0   & 0.0   & 0.9   & 0.0   & 0.7   & 2.1   & 4.5 \\
\cline{2-13}          & \multirow{4}[0]{*}{Microwave} & display & \textbf{25.0} & 0.0   & 12.9  & 0.0   & 0.0   & 0.4   & 0.0   & 3.3   & 14.5  & 5.2 \\
          &       & door  & 63.6  & \textbf{75.4} & 44.9  & 25.0  & 63.9  & 51.8  & 26.5  & 62.0  & 45.2  & 39.9 \\
          &       & handle & 73.1  & 86.6  & 84.8  & 0.0   & 0.0   & 33.2  & 0.0   & 37.7  & \textbf{95.2} & 0.0 \\
          &       & button & 12.5  & 0.0   & 10.4  & 0.0   & 0.0   & 5.3   & 0.0   & 4.8   & 15.9  & \textbf{21.3} \\
\cline{2-13}          & \multirow{2}[0]{*}{Refrigerator} & door  & 56.5  & \textbf{87.8} & 43.3  & 39.2  & 83.6  & 39.7  & 21.5  & 72.1  & 58.4  & 26.3 \\
          &       & handle & 30.3  & \textbf{64.5} & 50.4  & 0.0   & 0.0   & 31.0  & 0.0   & 13.6  & 53.1  & 14.1 \\
\cline{2-13}          & \multirow{3}[0]{*}{Scissors} & blade & 59.0  & 82.1  & \textbf{85.2} & 44.5  & 72.7  & 74.0  & 52.6  & 45.4  & 76.8  & 65.4 \\
          &       & handle & 78.1  & 89.8  & \textbf{90.8} & 65.2  & 83.4  & 79.0  & 64.7  & 79.7  & 86.8  & 0.0 \\
          &       & screw & 12.8  & 0.0   & \textbf{52.0} & 0.0   & 0.0   & 14.0  & 0.0   & 3.9   & 17.4  & 0.0 \\
\cline{2-13}          & \multirow{3}[0]{*}{StorageFurniture} & door  & 64.2  & \textbf{71.9} & 69.1  & 25.2  & 61.9  & 21.6  & 22.5  & 54.7  & 56.4  & 45.8 \\
          &       & drawer & 65.6  & \textbf{80.8} & 43.9  & 0.0   & 0.0   & 17.0  & 0.3   & 26.7  & 33.0  & 26.4 \\
          &       & handle & 10.9  & 52.8  & 67.6  & 0.0   & 0.0   & 18.0  & 0.0   & 9.2   & \textbf{71.4} & 16.2 \\
\cline{2-13}          & \multirow{6}[0]{*}{Table} & door  & \textbf{71.7} & 14.5  & 33.6  & 0.0   & 0.0   & 0.0   & 0.0   & 0.0   & 0.0   & 24.7 \\
          &       & drawer & 42.3  & \textbf{55.6} & 41.0  & 8.3   & 35.0  & 29.1  & 22.0  & 24.9  & 35.3  & 35.0 \\
          &       & leg   & 67.3  & \textbf{85.0} & 64.4  & 15.8  & 15.4  & 45.7  & 17.7  & 53.7  & 66.4  & 56.4 \\
          &       & tabletop & 80.2  & \textbf{93.8} & 74.7  & 19.7  & 82.2  & 55.0  & 41.1  & 74.5  & 79.7  & 77.7 \\
          &       & wheel & 80.0  & 51.8  & 58.9  & 0.0   & 0.0   & 0.0   & 0.0   & 0.0   & 61.0  & \textbf{87.1} \\
          &       & handle & 40.9  & 11.8  & \textbf{56.3} & 0.0   & 0.0   & 19.4  & 0.0   & 1.2   & 12.3  & 5.2 \\
\cline{2-13}          & \multirow{3}[0]{*}{TrashCan} & footpedal & \textbf{82.3} & 0.0   & 1.4   & 0.0   & 0.0   & 0.9   & 0.0   & 37.7  & 0.0   & 2.4 \\
          &       & lid   & 55.5  & \textbf{68.5} & 49.7  & 4.0   & 59.6  & 26.9  & 0.0   & 60.9  & 64.8  & 63.5 \\
          &       & door  & \textbf{77.4} & 0.0   & 0.0   & 0.9   & 0.0   & 0.0   & 0.0   & 0.0   & 2.1   & 24.5 \\
\cmidrule{2-13}          & \multicolumn{2}{c|}{Overall (17)} & 55.6  & \textbf{58.5} & 50.2  & 18.1  & 39.2  & 32.8  & 19.2  & 41.1  & \textbf{56.3} & 31.8 \\
    \bottomrule
    \end{tabular}%
    \vspace{-1.5em}
  \label{tab:full_sem_0}%
\end{table*}%

\begin{table*}[t]
  \centering
  \scriptsize
  \setlength{\tabcolsep}{2.5pt}
  \setlength\extrarowheight{1pt}
  \caption{Full table (2/2) of semantic segmentation results on the PartNetE dataset. Category mIoUs are shown. For 17 overlapping object categories, baseline models leverage additional 28k training shapes in the 45x8+28k setting. For the other 28 non-overlapping object categories, there are only 8 shapes per object category during training.}
    \begin{tabular}{ccc|rrr|rrrrrr|r}
    \toprule
    \multicolumn{1}{c|}{\multirow{56}[0]{*}{\begin{sideways}Non-Overlapping Categories (27)\end{sideways}}} & \multicolumn{2}{c|}{} & \multicolumn{3}{c|}{few-shot w/ additional data (45x8+28k)} & \multicolumn{6}{c|}{few-shot (45x8)} & \multicolumn{1}{c}{zero-shot} \\
\cmidrule{2-13}    \multicolumn{1}{c|}{} & category & part  & \multicolumn{1}{c}{PointNet++~\cite{qi2017pointnet++}} & \multicolumn{1}{c}{PointNext~\cite{qian2022pointnext}} & \multicolumn{1}{c|}{SoftGroup~\cite{vu2022softgroup}} & \multicolumn{1}{c}{PointNet++~\cite{qi2017pointnet++}} & \multicolumn{1}{c}{PointNext~\cite{qian2022pointnext}} & \multicolumn{1}{c}{SoftGroup~\cite{vu2022softgroup}} & \multicolumn{1}{c}{ACD~\cite{gadelha2020label}} & \multicolumn{1}{c}{Prototype~\cite{zhao2021few}} & \multicolumn{1}{c|}{Ours} & \multicolumn{1}{c}{Ours} \\
\cmidrule{2-13}    \multicolumn{1}{c|}{} & Box   & lid   & 18.6  & 84.2  & 8.8   & 24.5  & 69.4  & 24.1  & 21.1  & 68.8  & \textbf{84.5} & 57.5 \\
\cline{2-13}    \multicolumn{1}{c|}{} & Bucket & handle & 0.0   & 4.1   & 25.0  & 0.0   & 0.0   & 18.9  & 0.0   & 31.3  & \textbf{36.5} & 2.0 \\
\cline{2-13}    \multicolumn{1}{c|}{} & \multirow{2}[0]{*}{Camera} & button & 0.0   & 0.0   & 12.6  & 0.0   & 0.0   & 13.9  & 0.0   & 6.0   & \textbf{43.2} & 14.2 \\
    \multicolumn{1}{c|}{} &       & lens  & 13.0  & 66.4  & 34.6  & 19.4  & 51.9  & 43.3  & 20.2  & 58.0  & \textbf{73.4} & 28.6 \\
\cline{2-13}    \multicolumn{1}{c|}{} & Cart  & wheel & 6.4   & 36.3  & 23.9  & 11.6  & 47.7  & 40.8  & 31.5  & 36.8  & \textbf{88.1} & 87.7 \\
\cline{2-13}    \multicolumn{1}{c|}{} & \multirow{4}[0]{*}{CoffeeMachine} & button & \textbf{32.6} & 0.0   & 2.4   & 0.0   & 0.0   & 4.3   & 0.0   & 0.7   & 6.4   & 6.3 \\
    \multicolumn{1}{c|}{} &       & container & 29.0  & 25.8  & 4.6   & 7.6   & 23.0  & 25.5  & 2.8   & 25.9  & \textbf{51.1} & 27.3 \\
    \multicolumn{1}{c|}{} &       & knob  & \textbf{32.6} & 3.6   & 8.2   & 0.0   & 0.0   & 1.3   & 0.0   & 7.8   & 32.6  & 17.5 \\
    \multicolumn{1}{c|}{} &       & lid   & 44.0  & 42.3  & 17.8  & 11.2  & 45.0  & 27.6  & 0.0   & 45.7  & \textbf{61.2} & 50.3 \\
\cline{2-13}    \multicolumn{1}{c|}{} & \multirow{2}[0]{*}{Dispenser} & head  & 18.0  & 20.7  & 18.3  & 6.9   & 34.1  & 42.8  & 22.0  & 45.2  & \textbf{60.4} & 25.0 \\
    \multicolumn{1}{c|}{} &       & lid   & 6.1   & 31.2  & 19.5  & 7.0   & 11.0  & 43.0  & 16.7  & 61.6  & \textbf{87.1} & 7.9 \\
\cline{2-13}    \multicolumn{1}{c|}{} & \multirow{2}[0]{*}{Eyeglasses} & body  & 77.2  & 93.0  & 77.8  & 85.8  & \textbf{94.1} & 74.5  & 82.6  & 81.7  & 84.8  & 0.6 \\
    \multicolumn{1}{c|}{} &       & leg   & 75.1  & 83.2  & 67.0  & 71.8  & 84.6  & 70.9  & 73.7  & 74.0  & \textbf{91.7} & 3.0 \\
\cline{2-13}    \multicolumn{1}{c|}{} & FoldingChair & seat  & 10.9  & \textbf{96.4} & 14.7  & 63.4  & 94.9  & 89.0  & 74.2  & 91.2  & 86.3  & 91.7 \\
\cline{2-13}    \multicolumn{1}{c|}{} & Globe & sphere & 46.5  & 92.3  & 59.0  & 51.4  & 88.8  & 85.1  & 69.8  & 88.3  & \textbf{95.7} & 34.8 \\
\cline{2-13}    \multicolumn{1}{c|}{} & \multirow{3}[0]{*}{Kettle} & lid   & 16.2  & 24.5  & 46.9  & 21.4  & 54.7  & 60.2  & 22.9  & 58.9  & \textbf{78.8} & 30.9 \\
    \multicolumn{1}{c|}{} &       & handle & 16.2  & 71.3  & 56.8  & 33.8  & 73.1  & 60.1  & 43.7  & \textbf{73.6} & 73.5  & 31.4 \\
    \multicolumn{1}{c|}{} &       & spout & 30.2  & 39.6  & 68.5  & 30.5  & 53.7  & 61.8  & 54.0  & 55.5  & \textbf{78.6} & 0.0 \\
\cline{2-13}    \multicolumn{1}{c|}{} & \multirow{2}[0]{*}{KitchenPot} & lid   & 25.9  & 79.6  & 49.1  & 44.1  & \textbf{80.1} & 66.8  & 69.9  & 76.1  & 77.7  & 4.8 \\
    \multicolumn{1}{c|}{} &       & handle & 5.7   & 34.3  & 41.9  & 19.3  & 51.8  & 42.7  & 33.8  & 50.5  & \textbf{61.5} & 4.6 \\
\cline{2-13}    \multicolumn{1}{c|}{} & \multirow{3}[0]{*}{Lighter} & lid   & 52.4  & 38.4  & 32.0  & 33.6  & 39.9  & 40.5  & 32.3  & 42.8  & \textbf{69.9} & 69.1 \\
    \multicolumn{1}{c|}{} &       & wheel & 15.0  & 10.5  & 24.3  & 0.8   & 0.0   & 35.3  & 0.0   & 15.4  & \textbf{57.9} & 27.8 \\
    \multicolumn{1}{c|}{} &       & button & 37.6  & 0.0   & 34.2  & 0.0   & 0.0   & 43.7  & 0.0   & 34.0  & \textbf{66.3} & 9.3 \\
\cline{2-13}    \multicolumn{1}{c|}{} & \multirow{3}[0]{*}{Mouse} & button & 3.0   & 0.8   & \textbf{20.2} & 0.0   & 2.7   & 4.8   & 0.0   & 0.1   & 16.2  & 1.6 \\
    \multicolumn{1}{c|}{} &       & cord  & 33.3  & 65.0  & 41.0  & 0.0   & 0.0   & 53.2  & 0.0   & 40.7  & \textbf{66.5} & 65.4 \\
    \multicolumn{1}{c|}{} &       & wheel & 0.0   & 0.0   & \textbf{70.8} & 0.0   & 0.0   & 31.9  & 0.0   & 19.4  & 49.4  & 14.0 \\
\cline{2-13}    \multicolumn{1}{c|}{} & \multirow{2}[0]{*}{Oven} & door  & 32.3  & \textbf{75.6} & 17.2  & 38.9  & 73.5  & 49.7  & 17.8  & 68.3  & 73.1  & 66.1 \\
    \multicolumn{1}{c|}{} &       & knob  & 36.4  & 0.0   & 10.1  & 0.0   & 0.0   & 21.5  & 0.0   & 4.7   & \textbf{73.9} & 0.0 \\
\cline{2-13}    \multicolumn{1}{c|}{} & \multirow{2}[0]{*}{Pen} & cap   & 42.7  & 53.3  & 26.3  & 8.8   & 45.4  & 40.5  & 10.8  & 34.0  & \textbf{68.4} & 29.2 \\
    \multicolumn{1}{c|}{} &       & button & 50.3  & 25.6  & 31.4  & 0.0   & 21.0  & 52.1  & 0.0   & 61.0  & \textbf{74.6} & 0.0 \\
\cline{2-13}    \multicolumn{1}{c|}{} & \multirow{2}[0]{*}{Phone} & lid   & 40.0  & \textbf{78.7} & 0.3   & 10.3  & 66.7  & 2.0   & 19.7  & 68.3  & 74.0  & 48.5 \\
    \multicolumn{1}{c|}{} &       & button & 0.0   & 0.2   & 4.4   & 0.0   & 0.0   & 8.2   & 0.0   & 2.6   & 22.8  & \textbf{23.7} \\
\cline{2-13}    \multicolumn{1}{c|}{} & Pliers & leg   & 57.7  & \textbf{99.6} & 74.2  & 99.3  & \textbf{99.6} & 91.2  & 83.5  & 91.0  & 33.2  & 5.4 \\
\cline{2-13}    \multicolumn{1}{c|}{} & Printer & button & 0.0   & 0.0   & 1.2   & 0.0   & 0.0   & 1.6   & 0.0   & 0.2   & \textbf{4.3} & 0.8 \\
\cline{2-13}    \multicolumn{1}{c|}{} & Remote & button & 3.6   & \textbf{57.8} & 37.1  & 0.0   & 0.5   & 37.5  & 0.0   & 29.6  & 38.3  & 11.5 \\
\cline{2-13}    \multicolumn{1}{c|}{} & \multirow{3}[0]{*}{Safe} & door  & 14.0  & \textbf{76.7} & 9.8   & 32.7  & 67.0  & 24.8  & 28.0  & 51.9  & 64.5  & 34.5 \\
    \multicolumn{1}{c|}{} &       & switch & 13.6  & 0.0   & 5.8   & 0.0   & 0.0   & 21.7  & 0.0   & 5.8   & \textbf{27.9} & 4.3 \\
    \multicolumn{1}{c|}{} &       & button & \textbf{68.2} & 0.0   & 0.4   & 0.0   & 0.0   & 0.0   & 0.0   & 2.7   & 4.1   & 28.4 \\
\cline{2-13}    \multicolumn{1}{c|}{} & \multirow{2}[0]{*}{Stapler} & body  & 58.3  & 91.4  & 83.4  & 30.4  & 91.1  & 83.9  & 49.8  & 83.0  & \textbf{93.6} & 2.1 \\
    \multicolumn{1}{c|}{} &       & lid   & 44.9  & \textbf{85.7} & 76.8  & 45.7  & 83.3  & 80.5  & 50.2  & 78.4  & 76.0  & 39.6 \\
\cline{2-13}    \multicolumn{1}{c|}{} & \multirow{2}[0]{*}{Suitcase} & handle & 6.3   & 9.3   & 30.0  & 6.7   & 28.9  & 30.7  & 26.4  & 38.9  & \textbf{84.1} & 23.4 \\
    \multicolumn{1}{c|}{} &       & wheel & \textbf{75.0} & 17.8  & 6.6   & 0.0   & 0.0   & 28.9  & 0.0   & 32.1  & 56.7  & 57.0 \\
\cline{2-13}    \multicolumn{1}{c|}{} & Switch & switch & 1.8   & 39.7  & 21.0  & 9.3   & 42.9  & 31.8  & 10.3  & 40.9  & \textbf{59.4} & 9.5 \\
\cline{2-13}    \multicolumn{1}{c|}{} & \multirow{2}[0]{*}{Toaster} & button & 23.5  & 2.7   & 36.6  & 0.0   & 0.0   & 17.7  & 0.0   & 9.0   & \textbf{58.7} & 27.6 \\
    \multicolumn{1}{c|}{} &       & slider & 5.9   & 14.0  & 16.2  & 0.0   & 0.0   & 11.8  & 0.0   & 11.2  & \textbf{61.3} & 0.0 \\
\cline{2-13}    \multicolumn{1}{c|}{} & \multirow{3}[0]{*}{Toilet} & lid   & 19.5  & 49.4  & 12.7  & 9.4   & 68.5  & 27.9  & 53.4  & 56.8  & \textbf{72.6} & 35.0 \\
    \multicolumn{1}{c|}{} &       & seat  & \textbf{62.3} & 0.0   & 2.9   & 0.0   & 0.0   & 6.2   & 0.0   & 0.1   & 21.3  & 15.4 \\
    \multicolumn{1}{c|}{} &       & button & 16.4  & 0.0   & 23.2  & 0.0   & 0.0   & 7.6   & 0.0   & 1.6   & \textbf{67.6} & 11.4 \\
\cline{2-13}    \multicolumn{1}{c|}{} & \multirow{2}[0]{*}{USB} & cap   & 54.9  & 67.2  & 61.6  & 21.1  & \textbf{79.7} & 73.9  & 11.4  & 72.6  & 58.1  & 21.7 \\
    \multicolumn{1}{c|}{} &       & rotation & 49.8  & \textbf{68.6} & 26.6  & 35.7  & 61.7  & 38.1  & 38.9  & 58.1  & 50.7  & 0.0 \\
\cline{2-13}    \multicolumn{1}{c|}{} & \multirow{2}[0]{*}{WashingMachine} & door  & 1.1   & 54.5  & 25.8  & 8.9   & 37.9  & 40.0  & 20.2  & 55.4  & \textbf{63.3} & 19.3 \\
    \multicolumn{1}{c|}{} &       & button & 0.0   & 0.0   & 22.4  & 0.0   & 0.0   & 5.0   & 0.0   & 6.7   & \textbf{43.6} & 5.6 \\
\cline{2-13}    \multicolumn{1}{c|}{} & Window & window & 26.3  & \textbf{83.3} & 39.2  & 62.6  & 83.2  & 66.4  & 66.8  & 76.6  & 75.4  & 5.2 \\
\cmidrule{2-13}    \multicolumn{1}{c|}{} & \multicolumn{2}{c|}{Overall (28)} & 25.4  & \textbf{45.1}  & 30.7  & 21.8  & 41.5  & 41.1  & 25.6  & 46.3  & \textbf{61.3} & 24.4 \\
    \midrule
    \multicolumn{3}{c|}{Overall (45)} & 36.8  & \textbf{50.2}  & 38.1  & 20.4  & 40.6  & 38.0  & 23.2  & 44.3  & \textbf{59.4} & 27.2 \\
    \bottomrule
    \end{tabular}%
    \vspace{-1.5em}
  \label{tab:full_sem_1}%
\end{table*}%

\begin{table*}[t]
  \centering
  \scriptsize
  \setlength{\tabcolsep}{0.7pt}
  \setlength\extrarowheight{1pt}
  \caption{The full table of instance segmentation results on the PartNetE dataset. Category mAP50s ($\%$) are shown. For 17 overlapping object categories, baseline approaches leverage additional 28k training shapes in the 45x8+28k setting. For the other 28 non-overlapping object categories, there are only 8 shapes per object category during training.}
    \begin{tabular}{rrrrrrrrr|ccc|rr|rrr|r}
    \toprule
    \multicolumn{1}{c|}{\multirow{54}[0]{*}{\begin{sideways}Overlapping Categories\end{sideways}}} & \multicolumn{1}{c}{\multirow{3}[0]{*}{category}} & \multicolumn{1}{c|}{\multirow{3}[0]{*}{part}} & \multicolumn{2}{c|}{45x8+28k} & \multicolumn{3}{c|}{few-shot (45x8)} & \multicolumn{1}{c|}{zero-shot} & \multicolumn{1}{c|}{\multirow{57}[0]{*}{\begin{sideways}Non-Overlapping Categories\end{sideways}}} & \multirow{3}[0]{*}{category} & \multirow{3}[0]{*}{part} & \multicolumn{2}{c|}{45x8+28k} & \multicolumn{3}{c|}{few-shot (45x8)} & \multicolumn{1}{c}{zero-shot} \\
\cline{4-9}\cline{13-18}    \multicolumn{1}{c|}{} &       & \multicolumn{1}{c|}{} & \multicolumn{1}{c}{Point} & \multicolumn{1}{c|}{Soft} & \multicolumn{1}{c}{Point} & \multicolumn{1}{c}{Soft} & \multicolumn{1}{c|}{\multirow{2}[0]{*}{Ours}} & \multicolumn{1}{c|}{\multirow{2}[0]{*}{Ours}} & \multicolumn{1}{c|}{} &       &       & \multicolumn{1}{c}{Point} & \multicolumn{1}{c|}{Soft} & \multicolumn{1}{c}{Point} & \multicolumn{1}{c}{Soft} & \multicolumn{1}{c|}{\multirow{2}[0]{*}{Ours}} & \multicolumn{1}{c}{\multirow{2}[0]{*}{Ours}} \\
    \multicolumn{1}{c|}{} &       & \multicolumn{1}{c|}{} & \multicolumn{1}{c}{Group~\cite{jiang2020pointgroup}} & \multicolumn{1}{c|}{Group~\cite{vu2022softgroup}} & \multicolumn{1}{c}{Group~\cite{jiang2020pointgroup}} & \multicolumn{1}{c}{Group~\cite{vu2022softgroup}} & \multicolumn{1}{c|}{} &       & \multicolumn{1}{c|}{} &       &       & \multicolumn{1}{c}{Group~\cite{jiang2020pointgroup}} & \multicolumn{1}{c|}{Group~\cite{vu2022softgroup}} & \multicolumn{1}{c}{Group~\cite{jiang2020pointgroup}} & \multicolumn{1}{c}{Group~\cite{vu2022softgroup}} &       &  \\
\cline{2-9}\cline{11-18}    \multicolumn{1}{c|}{} & \multicolumn{1}{c}{Bottle} & \multicolumn{1}{c|}{lid} & 38.2  & \multicolumn{1}{r|}{43.9} & 8.0   & 22.4  & \multicolumn{1}{r|}{\textbf{79.4}} & 75.5  & \multicolumn{1}{c|}{} & Box   & lid   & 7.2   & 8.6   & 15.8  & 19.7  & \textbf{77.2} & 24.2 \\
\cline{2-9}\cline{11-18}    \multicolumn{1}{c|}{} & \multicolumn{1}{c}{\multirow{5}[0]{*}{Chair}} & \multicolumn{1}{c|}{arm} & 94.6  & \multicolumn{1}{r|}{\textbf{95.1}} & 35.9  & 71.0  & \multicolumn{1}{r|}{67.7} & 23.9  & \multicolumn{1}{c|}{} & Bucket & handle & 1.5   & 1.6   & 1.0   & 1.1   & \textbf{18.2} & 5.9 \\
\cline{11-18}    \multicolumn{1}{c|}{} &       & \multicolumn{1}{c|}{back} & 82.0  & \multicolumn{1}{r|}{73.2} & 83.8  & 93.7  & \multicolumn{1}{r|}{\textbf{95.4}} & 30.0  & \multicolumn{1}{c|}{} & \multirow{2}[0]{*}{Camera} & button & 1.0   & 1.5   & 4.5   & 6.1   & \textbf{33.8} & 11.9 \\
    \multicolumn{1}{c|}{} &       & \multicolumn{1}{c|}{leg} & 88.6  & \multicolumn{1}{r|}{\textbf{93.6}} & 92.2  & 89.9  & \multicolumn{1}{r|}{78.1} & 30.3  & \multicolumn{1}{c|}{} &       & lens  & 16.1  & 0.0   & 5.0   & 16.4  & \textbf{39.9} & 4.9 \\
\cline{11-18}    \multicolumn{1}{c|}{} &       & \multicolumn{1}{c|}{seat} & 75.0  & \multicolumn{1}{r|}{85.9} & 81.4  & 88.1  & \multicolumn{1}{r|}{85.5} & \textbf{88.9} & \multicolumn{1}{c|}{} & Cart  & wheel & 29.2  & 28.4  & 28.5  & 29.8  & \textbf{83.3} & 79.3 \\
\cline{11-18}    \multicolumn{1}{c|}{} &       & \multicolumn{1}{c|}{wheel} & 98.0  & \multicolumn{1}{r|}{97.7} & 92.8  & 95.9  & \multicolumn{1}{r|}{95.5} & \textbf{99.3} & \multicolumn{1}{c|}{} & \multirow{4}[0]{*}{CoffeeMachine} & button & 1.0   & 1.0   & 1.1   & 0.0   & \textbf{2.2} & 1.8 \\
\cline{2-9}    \multicolumn{1}{c|}{} & \multicolumn{1}{c}{Clock} & \multicolumn{1}{c|}{hand} & 1.0   & \multicolumn{1}{r|}{1.0} & 1.0   & 1.0   & \multicolumn{1}{r|}{\textbf{14.9}} & 4.2   & \multicolumn{1}{c|}{} &       & container & 2.5   & 4.0   & 13.6  & 19.7  & \textbf{32.8} & 7.1 \\
\cline{2-9}    \multicolumn{1}{c|}{} & \multicolumn{1}{c}{\multirow{2}[0]{*}{Dishwasher}} & \multicolumn{1}{c|}{door} & \textbf{76.7} & \multicolumn{1}{r|}{75.0} & 50.6  & 55.6  & \multicolumn{1}{r|}{57.4} & 22.5  & \multicolumn{1}{c|}{} &       & knob  & 5.6   & 5.0   & 3.3   & 1.5   & \textbf{13.5} & 7.2 \\
    \multicolumn{1}{c|}{} &       & \multicolumn{1}{c|}{handle} & 55.6  & \multicolumn{1}{r|}{\textbf{56.4}} & 1.0   & 26.4  & \multicolumn{1}{r|}{32.9} & 0.0   & \multicolumn{1}{c|}{} &       & lid   & 3.3   & 1.4   & 8.9   & 22.6  & \textbf{27.6} & 19.5 \\
\cline{2-9}\cline{11-18}    \multicolumn{1}{c|}{} & \multicolumn{1}{c}{\multirow{3}[0]{*}{Display}} & \multicolumn{1}{c|}{base} & 95.2  & \multicolumn{1}{r|}{\textbf{97.4}} & 13.2  & 22.1  & \multicolumn{1}{r|}{94.2} & 58.3  & \multicolumn{1}{c|}{} & \multirow{2}[0]{*}{Dispenser} & head  & 27.5  & 29.2  & 39.1  & 45.4  & \textbf{46.4} & 13.7 \\
    \multicolumn{1}{c|}{} &       & \multicolumn{1}{c|}{screen} & 46.0  & \multicolumn{1}{r|}{55.4} & 32.9  & 49.2  & \multicolumn{1}{r|}{\textbf{70.7}} & 40.5  & \multicolumn{1}{c|}{} &       & lid   & 20.5  & 23.6  & 22.4  & 30.2  & \textbf{80.6} & 5.0 \\
\cline{11-18}    \multicolumn{1}{c|}{} &       & \multicolumn{1}{c|}{support} & 54.0  & \multicolumn{1}{r|}{53.2} & 4.1   & 11.1  & \multicolumn{1}{r|}{\textbf{84.0}} & 0.0   & \multicolumn{1}{c|}{} & \multirow{2}[0]{*}{Eyeglasses} & body  & 31.7  & 39.5  & 28.1  & 34.7  & \textbf{79.5} & 1.0 \\
\cline{2-9}    \multicolumn{1}{c|}{} & \multicolumn{1}{c}{\multirow{3}[0]{*}{Door}} & \multicolumn{1}{c|}{frame} & \textbf{36.8} & \multicolumn{1}{r|}{28.3} & 2.7   & 9.8   & \multicolumn{1}{r|}{2.8} & 1.0   & \multicolumn{1}{c|}{} &       & leg   & 68.0  & 62.7  & 50.3  & 56.3  & \textbf{84.9} & 1.2 \\
\cline{11-18}    \multicolumn{1}{c|}{} &       & \multicolumn{1}{c|}{door} & 32.4  & \multicolumn{1}{r|}{\textbf{34.3}} & 7.5   & 5.9   & \multicolumn{1}{r|}{30.7} & 3.0   & \multicolumn{1}{c|}{} & FoldingChair & seat  & 16.8  & 16.8  & 86.4  & 79.0  & 76.7  & \textbf{87.0} \\
\cline{11-18}    \multicolumn{1}{c|}{} &       & \multicolumn{1}{c|}{handle} & 1.0   & \multicolumn{1}{r|}{1.0} & 1.0   & 1.0   & \multicolumn{1}{r|}{\textbf{20.3}} & 0.0   & \multicolumn{1}{c|}{} & Globe & sphere & 63.1  & 63.1  & 80.2  & 75.7  & \textbf{81.0} & 18.3 \\
\cline{2-9}\cline{11-18}    \multicolumn{1}{c|}{} & \multicolumn{1}{c}{\multirow{2}[0]{*}{Faucet}} & \multicolumn{1}{c|}{spout} & 85.4  & \multicolumn{1}{r|}{\textbf{86.3}} & 50.7  & 52.4  & \multicolumn{1}{r|}{61.7} & 3.1   & \multicolumn{1}{c|}{} & \multirow{3}[0]{*}{Kettle} & lid   & 64.0  & 64.4  & 65.8  & 70.0  & \textbf{76.1} & 30.9 \\
    \multicolumn{1}{c|}{} &       & \multicolumn{1}{c|}{switch} & \textbf{74.5} & \multicolumn{1}{r|}{72.5} & 11.2  & 22.2  & \multicolumn{1}{r|}{47.6} & 1.5   & \multicolumn{1}{c|}{} &       & handle & 51.4  & 54.3  & 45.0  & 59.0  & \textbf{78.1} & 22.9 \\
\cline{2-9}    \multicolumn{1}{c|}{} & \multicolumn{1}{c}{\multirow{2}[0]{*}{Keyboard}} & \multicolumn{1}{c|}{cord} & 42.6  & \multicolumn{1}{r|}{39.7} & 34.3  & 21.3  & \multicolumn{1}{r|}{\textbf{68.6}} & 25.0  & \multicolumn{1}{c|}{} &       & spout & 68.5  & \textbf{72.6} & 45.4  & 61.3  & 71.9  & 1.0 \\
\cline{11-18}    \multicolumn{1}{c|}{} &       & \multicolumn{1}{c|}{key} & 37.2  & \multicolumn{1}{r|}{\textbf{37.7}} & 16.1  & 1.0   & \multicolumn{1}{r|}{12.3} & 1.0   & \multicolumn{1}{c|}{} & \multirow{2}[0]{*}{KitchenPot} & lid   & 68.3  & 68.5  & 81.4  & 87.1  & \textbf{91.5} & 1.0 \\
\cline{2-9}    \multicolumn{1}{c|}{} & \multicolumn{1}{c}{Knife} & \multicolumn{1}{c|}{blade} & 19.3  & \multicolumn{1}{r|}{27.2} & 15.6  & 10.3  & \multicolumn{1}{r|}{\textbf{43.9}} & 22.1  & \multicolumn{1}{c|}{} &       & handle & \textbf{50.6} & 50.1  & 32.5  & 44.3  & 49.5  & 1.3 \\
\cline{2-9}\cline{11-18}    \multicolumn{1}{c|}{} & \multicolumn{1}{c}{\multirow{4}[0]{*}{Lamp}} & \multicolumn{1}{c|}{base} & 64.3  & \multicolumn{1}{r|}{71.1} & 8.5   & 17.9  & \multicolumn{1}{r|}{\textbf{89.9}} & 87.2  & \multicolumn{1}{c|}{} & \multirow{3}[0]{*}{Lighter} & lid   & 30.7  & 30.7  & 0.0   & 40.6  & \textbf{45.8} & 24.1 \\
    \multicolumn{1}{c|}{} &       & \multicolumn{1}{c|}{body} & 48.6  & \multicolumn{1}{r|}{36.5} & 4.3   & 11.0  & \multicolumn{1}{r|}{\textbf{87.4}} & 1.0   & \multicolumn{1}{c|}{} &       & wheel & 6.0   & 5.3   & 0.0   & \textbf{47.9} & 34.3  & 16.6 \\
    \multicolumn{1}{c|}{} &       & \multicolumn{1}{c|}{bulb} & 54.5  & \multicolumn{1}{r|}{\textbf{59.2}} & 7.1   & 1.9   & \multicolumn{1}{r|}{5.9} & 5.9   & \multicolumn{1}{c|}{} &       & button & 64.1  & \textbf{67.8} & 0.0   & 63.2  & 23.6  & 1.8 \\
\cline{11-18}    \multicolumn{1}{c|}{} &       & \multicolumn{1}{c|}{shade} & 83.5  & \multicolumn{1}{r|}{86.4} & 19.4  & 47.0  & \multicolumn{1}{r|}{\textbf{90.1}} & 49.0  & \multicolumn{1}{c|}{} & \multirow{3}[0]{*}{Mouse} & button & 1.0   & 1.0   & 0.0   & 0.0   & \textbf{1.7} & \textbf{1.7} \\
\cline{2-9}    \multicolumn{1}{c|}{} & \multicolumn{1}{c}{\multirow{5}[0]{*}{Laptop}} & \multicolumn{1}{c|}{keyboard} & 0.0   & \multicolumn{1}{r|}{0.0} & 40.1  & \textbf{53.8} & \multicolumn{1}{r|}{53.4} & 42.5  & \multicolumn{1}{c|}{} &       & cord  & 1.0   & 1.0   & 0.0   & 1.0   & \textbf{66.3} & \textbf{66.3} \\
    \multicolumn{1}{c|}{} &       & \multicolumn{1}{c|}{screen} & 1.0   & \multicolumn{1}{r|}{1.0} & 36.3  & \textbf{61.5} & \multicolumn{1}{r|}{48.5} & 59.5  & \multicolumn{1}{c|}{} &       & wheel & \textbf{83.2} & \textbf{83.2} & 0.0   & 53.7  & 50.5  & 8.9 \\
\cline{11-18}    \multicolumn{1}{c|}{} &       & \multicolumn{1}{c|}{shaft} & 1.2   & \multicolumn{1}{r|}{\textbf{3.5}} & 1.0   & 0.0   & \multicolumn{1}{r|}{2.0} & 0.0   & \multicolumn{1}{c|}{} & \multirow{2}[0]{*}{Oven} & door  & 26.5  & 31.9  & 0.0   & 19.1  & \textbf{54.9} & 36.4 \\
    \multicolumn{1}{c|}{} &       & \multicolumn{1}{c|}{touchpad} & 0.0   & \multicolumn{1}{r|}{0.0} & 0.0   & 0.0   & \multicolumn{1}{r|}{\textbf{19.7}} & 9.9   & \multicolumn{1}{c|}{} &       & knob  & 1.0   & 1.0   & 0.0   & 1.6   & \textbf{74.1} & 15.4 \\
\cline{11-18}    \multicolumn{1}{c|}{} &       & \multicolumn{1}{c|}{camera} & 0.0   & \multicolumn{1}{r|}{0.0} & 0.0   & 0.0   & \multicolumn{1}{r|}{\textbf{1.0}} & 0.0   & \multicolumn{1}{c|}{} & \multirow{2}[0]{*}{Pen} & cap   & 48.2  & 44.4  & 0.0   & 44.3  & \textbf{51.6} & 7.8 \\
\cline{2-9}    \multicolumn{1}{c|}{} & \multicolumn{1}{c}{\multirow{4}[0]{*}{Microwave}} & \multicolumn{1}{c|}{display} & 4.2   & \multicolumn{1}{r|}{1.0} & 0.0   & 1.0   & \multicolumn{1}{r|}{\textbf{6.3}} & 1.0   & \multicolumn{1}{c|}{} &       & button & 16.9  & 16.9  & 0.0   & 10.9  & \textbf{37.9} & 1.0 \\
\cline{11-18}    \multicolumn{1}{c|}{} &       & \multicolumn{1}{c|}{door} & \textbf{62.6} & \multicolumn{1}{r|}{57.1} & 0.0   & 31.0  & \multicolumn{1}{r|}{34.4} & 19.3  & \multicolumn{1}{c|}{} & \multirow{2}[0]{*}{Phone} & lid   & 1.0   & 1.1   & 0.0   & 1.2   & \textbf{37.8} & 12.0 \\
    \multicolumn{1}{c|}{} &       & \multicolumn{1}{c|}{handle} & 1.0   & \multicolumn{1}{r|}{1.0} & 0.0   & 0.0   & \multicolumn{1}{r|}{\textbf{60.4}} & 0.0   & \multicolumn{1}{c|}{} &       & button & 1.0   & 1.0   & 0.0   & 1.0   & \textbf{26.6} & 2.8 \\
\cline{11-18}    \multicolumn{1}{c|}{} &       & \multicolumn{1}{c|}{button} & \textbf{100.0} & \multicolumn{1}{r|}{\textbf{100.0}} & 0.0   & 22.8  & \multicolumn{1}{r|}{3.2} & 4.0   & \multicolumn{1}{c|}{} & Pliers & leg   & 28.2  & \textbf{40.4} & 6.8   & 14.5  & 4.7   & 5.9 \\
\cline{2-9}\cline{11-18}    \multicolumn{1}{c|}{} & \multicolumn{1}{c}{\multirow{2}[0]{*}{Refrigerator}} & \multicolumn{1}{c|}{door} & \textbf{57.1} & \multicolumn{1}{r|}{54.2} & 0.0   & 23.2  & \multicolumn{1}{r|}{31.3} & 14.3  & \multicolumn{1}{c|}{} & Printer & button & 1.0   & 1.0   & 0.0   & 0.0   & \textbf{1.3} & 1.0 \\
\cline{11-18}    \multicolumn{1}{c|}{} &       & \multicolumn{1}{c|}{handle} & 19.3  & \multicolumn{1}{r|}{17.2} & 0.0   & 9.7   & \multicolumn{1}{r|}{\textbf{39.7}} & 8.6   & \multicolumn{1}{c|}{} & Remote & button & \textbf{23.4} & 22.5  & 0.0   & 6.2   & 23.1  & 3.5 \\
\cline{2-9}\cline{11-18}    \multicolumn{1}{c|}{} & \multicolumn{1}{c}{\multirow{3}[0]{*}{Scissors}} & \multicolumn{1}{c|}{blade} & 6.2   & \multicolumn{1}{r|}{6.5} & 4.5   & 3.0   & \multicolumn{1}{r|}{\textbf{14.1}} & 4.2   & \multicolumn{1}{c|}{} & \multirow{3}[0]{*}{Safe} & door  & 11.0  & 12.3  & 0.0   & 19.4  & \textbf{68.4} & 28.7 \\
    \multicolumn{1}{c|}{} &       & \multicolumn{1}{c|}{handle} & 82.0  & \multicolumn{1}{r|}{\textbf{82.9}} & 41.9  & 34.5  & \multicolumn{1}{r|}{58.4} & 0.0   & \multicolumn{1}{c|}{} &       & switch & 4.8   & 5.4   & 0.0   & 23.3  & \textbf{27.4} & 3.3 \\
    \multicolumn{1}{c|}{} &       & \multicolumn{1}{c|}{screw} & 27.2  & \multicolumn{1}{r|}{\textbf{28.4}} & 8.9   & 4.6   & \multicolumn{1}{r|}{4.3} & 0.0   & \multicolumn{1}{c|}{} &       & button & \textbf{1.0} & \textbf{1.0} & 0.0   & \textbf{1.0} & \textbf{1.0} & \textbf{1.0} \\
\cline{2-9}\cline{11-18}    \multicolumn{1}{c|}{} & \multicolumn{1}{c}{\multirow{3}[0]{*}{StorageFurniture}} & \multicolumn{1}{c|}{door} & \textbf{86.9} & \multicolumn{1}{r|}{85.6} & 0.0   & 28.8  & \multicolumn{1}{r|}{24.9} & 13.5  & \multicolumn{1}{c|}{} & \multirow{2}[0]{*}{Stapler} & body  & 86.6  & 96.7  & 52.4  & 88.0  & \textbf{100.0} & 1.0 \\
    \multicolumn{1}{c|}{} &       & \multicolumn{1}{c|}{drawer} & 3.9   & \multicolumn{1}{r|}{4.2} & 0.0   & 1.5   & \multicolumn{1}{r|}{6.1} & \textbf{8.0} & \multicolumn{1}{c|}{} &       & lid   & 90.0  & \textbf{91.8} & 69.8  & 78.2  & 89.7  & 36.0 \\
\cline{11-18}    \multicolumn{1}{c|}{} &       & \multicolumn{1}{c|}{handle} & 56.4  & \multicolumn{1}{r|}{57.5} & 0.0   & 4.6   & \multicolumn{1}{r|}{\textbf{67.5}} & 11.2  & \multicolumn{1}{c|}{} & \multirow{2}[0]{*}{Suitcase} & handle & 25.5  & 24.2  & 0.0   & 12.9  & \textbf{64.1} & 40.8 \\
\cline{2-9}    \multicolumn{1}{c|}{} & \multicolumn{1}{c}{\multirow{6}[0]{*}{Table}} & \multicolumn{1}{c|}{door} & 44.4  & \multicolumn{1}{r|}{\textbf{49.3}} & 0.0   & 0.0   & \multicolumn{1}{r|}{0.0} & 8.2   & \multicolumn{1}{c|}{} &       & wheel & 5.7   & 2.9   & 0.0   & 3.1   & 25.7  & \textbf{27.5} \\
\cline{11-18}    \multicolumn{1}{c|}{} &       & \multicolumn{1}{c|}{drawer} & 35.7  & \multicolumn{1}{r|}{\textbf{36.5}} & 0.0   & 0.0   & \multicolumn{1}{r|}{11.3} & 8.9   & \multicolumn{1}{c|}{} & Switch & switch & 7.5   & 5.6   & 0.0   & 21.2  & \textbf{35.1} & 5.6 \\
\cline{11-18}    \multicolumn{1}{c|}{} &       & \multicolumn{1}{c|}{leg} & 33.8  & \multicolumn{1}{r|}{27.4} & 0.0   & 7.7   & \multicolumn{1}{r|}{\textbf{45.9}} & 38.7  & \multicolumn{1}{c|}{} & \multirow{2}[0]{*}{Toaster} & button & 9.0   & 10.1  & 0.0   & 4.5   & \textbf{31.4} & 9.0 \\
    \multicolumn{1}{c|}{} &       & \multicolumn{1}{c|}{tabletop} & 81.2  & \multicolumn{1}{r|}{\textbf{82.0}} & 0.0   & 30.0  & \multicolumn{1}{r|}{64.1} & 65.7  & \multicolumn{1}{c|}{} &       & slider & 5.0   & 5.0   & 0.0   & 16.9  & \textbf{45.4} & 0.0 \\
\cline{11-18}    \multicolumn{1}{c|}{} &       & \multicolumn{1}{c|}{wheel} & 1.0   & \multicolumn{1}{r|}{1.3} & 0.0   & 1.1   & \multicolumn{1}{r|}{64.7} & \textbf{92.6} & \multicolumn{1}{c|}{} & \multirow{3}[0]{*}{Toilet} & lid   & 5.5   & 6.1   & 0.0   & 37.5  & \textbf{62.3} & 11.0 \\
    \multicolumn{1}{c|}{} &       & \multicolumn{1}{c|}{handle} & \textbf{81.9} & \multicolumn{1}{r|}{80.8} & 0.0   & 46.4  & \multicolumn{1}{r|}{7.6} & 5.5   & \multicolumn{1}{c|}{} &       & seat  & 0.0   & 0.0   & 0.0   & 1.0   & \textbf{4.2} & 1.9 \\
\cline{2-9}    \multicolumn{1}{c|}{} & \multicolumn{1}{c}{\multirow{3}[0]{*}{TrashCan}} & \multicolumn{1}{c|}{footpedal} & 34.8  & \multicolumn{1}{r|}{\textbf{35.3}} & 0.0   & 15.3  & \multicolumn{1}{r|}{0.0} & 2.3   & \multicolumn{1}{c|}{} &       & button & 1.0   & 1.0   & 0.0   & 1.5   & \textbf{70.3} & 18.8 \\
\cline{11-18}    \multicolumn{1}{c|}{} &       & \multicolumn{1}{c|}{lid} & 0.0   & \multicolumn{1}{r|}{0.0} & 0.0   & 1.0   & \multicolumn{1}{r|}{37.8} & \textbf{38.9} & \multicolumn{1}{c|}{} & \multirow{2}[0]{*}{USB} & cap   & 67.3  & \textbf{75.7} & 0.0   & 69.0  & 26.0  & 23.4 \\
    \multicolumn{1}{c|}{} &       & \multicolumn{1}{c|}{door} & 0.0   & \multicolumn{1}{r|}{0.0} & 0.0   & 1.0   & \multicolumn{1}{r|}{1.0} & \textbf{1.8} & \multicolumn{1}{c|}{} &       & rotation & 16.3  & 15.0  & 0.0   & \textbf{33.3} & 29.7  & 0.0 \\
\cline{2-9}\cline{11-18}    \multicolumn{1}{c|}{} & \multicolumn{2}{c|}{Overall (17)} & 41.7  & \multicolumn{1}{r|}{\textbf{42.4}} & 14.6  & 21.3  & \multicolumn{1}{r|}{\textbf{42.5}} & 20.9  & \multicolumn{1}{c|}{} & \multirow{2}[0]{*}{WashingMachine} & door  & 25.0  & 34.3  & 0.0   & 41.5  & \textbf{46.4} & 10.9 \\
\cline{1-9}          &       &       &       &       &       &       &       &       & \multicolumn{1}{c|}{} &       & button & 0.0   & 0.0   & 0.0   & 1.0   & \textbf{14.1} & 3.0 \\
\cline{11-18}          &       &       &       &       &       &       &       &       & \multicolumn{1}{c|}{} & Window & window & 21.2  & \textbf{26.4} & 0.0   & 4.3   & 15.6  & 1.3 \\
\cline{11-18}          &       &       &       &       &       &       &       &       & \multicolumn{1}{c|}{} & \multicolumn{2}{c|}{Overall (28)} & 24.6  & \textbf{25.6} & 16.8  & 28.4  & \textbf{46.2} & 16.2 \\
\cline{10-18}          &       &       &       &       &       &       &       &       & \multicolumn{3}{c|}{Overall (45)} & 31.0  & \textbf{31.9} & 16.0  & 25.7  & \textbf{44.8} & 18.0 \\
\bottomrule  
\end{tabular}%
\vspace{-1.5em}
  \label{tab:ins_full}%
\end{table*}%

\subsection{Details of Baselines}
\label{sec:baselines}
We train baseline approaches on our PartNetE dataset. 

\vspace{-1em}
\paragraph{PointNet++ and PointNext}
We use PointNext's official code base to train PointNet++ and PointNext for semantic segmentation under both the ``45x8'' and ``45x8+28k'' settings, as described in Section~\ref{sec:settings_baselines}. Specifically, we adapt the configurations\footnote{PointNext: \url{https://github.com/guochengqian/PointNeXt/blob/master/cfgs/shapenetpart/pointnext-s.yaml}, PointNet++: \url{https://github.com/guochengqian/PointNeXt/tree/master/cfgs/scannet/pointnet++_original.yaml}} provided by PointNext and randomly sample 10,000 points per shape as the network input. We use 148-class segmentation heads for both baselines, including 103 part classes and 45 background classes (one for each object category). For PointNext, we utilize a c32 model and take point positions, normals and heights as input. For PointNet++, the model takes point positions and normals as input. 

\vspace{-1em}
\paragraph{PointGroup and SoftGroup}
We use SoftGroup's official code base to train PointGroup and SoftGroup for instance segmentation under both the ``45x8'' and ``45x8+28k'' settings, as described in Section~\ref{sec:settings_baselines}. Specifically, the training includes two stages: 1) training a backbone module for semantic and offset prediction; 2) training the rest modules while freezing the backbone from stage 1. We randomly sample (up to) 50k points for each shape and utilize the point positions and normals as the network input.

For the first stage, there are 104 classes (including 103 part classes and one background class), and points are highly unbalanced across the classes. To avoid losses being dominated by several common part classes, we apply frequency-based class weights, calculated as the inverse square root of point frequency~\cite{mahajan2018exploring}, to cross-entropy and offset losses. We also disable data augmentations (e.g., elastic transform) designed for scene-scale datasets. The voxel scale for voxelization is set to 100, and the backbone network is initialized with pretrained checkpoint \texttt{hais\_ckpt\_spconv2.pth}. We train the backbone for 200 epochs with a batch size of 16. We apply cosine learning rate attenuation starting from epoch 45 with an initial learning rate of 0.001. 

In the second stage, we train the remaining modules for instance segmentation, while freezing the trained backbone from the first stage. We train the networks with a batch size of 4 and an initial learning rate of 0.004. Since the original code is evaluated on indoor segmentation, we empirically tuned the parameters. Specifically, for the ``45x8'' setting, the grouping radius, mean active, and classification score threshold are set to 0.02,  50, and 0.001, respectively. For the ``45x8+28k'' setting, the grouping radius, mean active, and classification score threshold are set to 0.01, 300, and 0.01, respectively. In the ``45x8+28k'' setting, the few-shot shapes are repeated 50 times in each epoch to mitigate the unbalanced data issue. The PointGroup is trained using a similar pipeline to SoftGroup, except using one-hot semantic results from the first-stage backbone instead of softmax results.

\vspace{-1em}
\paragraph{ACD} Inspired by \cite{gadelha2020label}, we utilize an auxiliary self-supervised task to aid few-shot learning. Specifically, we use CoACD~\cite{wei2022approximate} to decompose the mesh of each 3D shape into approximate convex components using a concavity threshold of 0.05, which results in a median of 18 components per shape. Using the decomposition results, we add an auxiliary contrastive loss to the pipeline of PointNet++ as \cite{gadelha2020label}. As a result, the network is trained with both contrastive and original segmentation losses. The auxiliary contrastive loss encourages points within each convex component to have similar features, while points in different components have different features. To compute the contrastive loss efficiently, we randomly sample 2.5k out of 10k points when calculating pairwise contrastive losses.
 
\vspace{-1em}
\paragraph{Prototype} Inspired by \cite{zhao2021few}, we also utilize prototype learning to build a few-shot baseline. Specifically, we construct prototype features using the learned point features (by the PointNext backbone, 96 dim) of 360 few-shot shapes. For each part category, we first sample up to 100 point features as the seed features using the furthest point sampling (FPS) in the feature space. We then group the point features into clusters according to their distances to the seed features. We take the average point features of each group to serve as prototype features, which results in 100 prototype features for each part category. For each test shape, we classify each point by finding the nearest prototype features. Note that we only consider prototype features of parts that the object category may have.

\subsection{Full Table of Quantitative Comparison}
\label{sec:full_table}

Table~\ref{tab:full_sem_0} and~\ref{tab:full_sem_1} show the full tables of semantic segmentation results (corresponding to Table~\ref{table:semseg}). Table~\ref{tab:ins_full} shows the full table of instance segmentation results (corresponding to Table~\ref{table:inseg}).

\newpage  

{\footnotesize
\bibliographystyle{ieee_fullname}
\bibliography{egbib}
}

\end{document}